\documentclass[10pt,journal,compsoc]{IEEEtran}

\usepackage{stmaryrd}
\usepackage{amsfonts}
\usepackage{graphicx}
\usepackage{graphics}
\usepackage{subfigure}
\usepackage{algorithm}
\usepackage{algorithmic}
\usepackage{amsmath}
\usepackage{amssymb}
\usepackage{multirow,booktabs}
\usepackage{color}
\usepackage{cite}
\usepackage{multirow}
\usepackage{ulem}
\usepackage{enumerate}
\usepackage{times}
\ifCLASSINFOpdf
\else
\fi
\begin{document}
%
\title{A Cooperative Framework for \\Fireworks Algorithm}
%
%
%
%

\author{Shaoqiu Zheng,~\IEEEmembership{Student Member,~IEEE,}
       Junzhi~Li,~\IEEEmembership{Student Member,~IEEE,}
       Andreas Janecek,\\
        and Ying Tan,~\IEEEmembership{Senior Member,~IEEE}
\IEEEcompsocitemizethanks{\IEEEcompsocthanksitem Shaoqiu Zheng, Junzhi Li and Ying Tan are with the Department of Machine Intelligence, School of Electronics Engineering and Computer Science, Peking University, Key Laboratory of Machine Perception (Ministry of Education), Peking University, Beijing,100871, P.R. China. (email: \{zhengshaoqiu, ljz, ytan\}@pku.edu.cn).
\IEEEcompsocthanksitem Andreas Janecek is with University of Vienna, Research Group Theory and Applications of Algorithms, 1090 Vienna, Austria
(email: andreas.janecek@univie.ac.at )
\IEEEcompsocthanksitem Ying Tan is the corresponding author.}
\thanks{Manuscript}}

\IEEEtitleabstractindextext{%
\begin{abstract}
This paper presents a cooperative framework for fireworks algorithm (CoFFWA).
A detailed analysis of existing fireworks algorithm (FWA) and its recently developed variants has revealed that ($i$) the selection strategy lead to
the contribution of the \textit{firework with the best fitness} (core firework) for the optimization overwhelms the contributions of \textit{the rest of fireworks} (non-core fireworks) in the explosion operator, ($ii$) the Gaussian mutation operator is not as effective as it is designed to be.
To overcome these limitations,
the CoFFWA is proposed, which can greatly enhance the exploitation ability of non-core fireworks by using independent selection operator and increase the exploration capacity by crowdness-avoiding cooperative strategy among the fireworks.
Experimental results on the CEC2013 benchmark functions suggest that CoFFWA outperforms the state-of-the-art FWA variants, artificial
bee colony, differential evolution,  the standard particle swarm optimization (SPSO) in 2007 and the most recent SPSO in 2011 in term of convergence performance.
\end{abstract}

}

\maketitle

\IEEEdisplaynontitleabstractindextext

%
\IEEEpeerreviewmaketitle

\ifCLASSOPTIONcompsoc
\IEEEraisesectionheading{\section{Introduction}\label{sec:introduction}}
\else
\section{Introduction}
\label{sec:introduction}
\fi
In the past two decades, many stochastic and population-based swarm intelligence (SI) algorithms have been proposed. These algorithms have shown great success in dealing with optimization problems in various application fields.
SI refers to the ability presented by the swarm behavior to solve problems through the direct or indirect interaction among the agents with their environment~\cite{engelbrecht2005fundamentals}.
Usually, SI system consists of a population of decentralized and simple natural or artificial agents, the iteration among the agents lead to the emergence of intelligent ability~\cite{beni1993swarm}.
By the observing and modeling of the cooperative swarm behavior of living creatures, artificial systems and the physical properties of non-living objects, researchers are trying to understand those mechanisms and use them to design new algorithms.

The famous SI algorithms comprise
ant colony optimization (ACO) by modeling the cooperative behavior among ant colony through environment to search for the shortest path from the colony to the food~\cite{dorigo2006ant},
and particle swarm optimization (PSO) by mimicking the aggregating motion of a flock of birds for searching for food ~\cite{bratton2007}.

Recently, the great successes of ACO and PSO for solving optimization problems greatly push the SI developments forward.
Algorithms inspired by collective biologic behavior of bee, glowworm, fish school, firefly, cuckoo, krill herd
\cite{karaboga2010artificial}\cite{krishnanand2009}\cite{filho2009}\cite{lukasik2009firefly}\cite{yang2010firefly}\cite{yang2005engineering}\cite{yang2009cuckoo}\cite{gandomi2012krill}
\cite{shi2011brain}\cite{blomeke2007bacterial}\cite{yang2010batAlgorithm}\cite{karaboga2005idea} and collective artificial system are proposed
~\cite{magnetic2008}~\cite{shah2009intelligent}\cite{tan2010fireworks}.
See \cite{xing2014emerging} for a detailed review of recently developed computational intelligence algorithms.

\begin{figure}
\centering
\subfigure[fireworks explosion]{
  \includegraphics[width=0.2\textwidth]{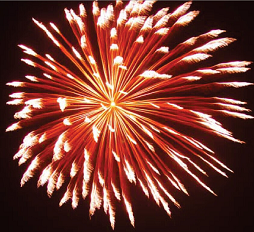}
  \label{Fig:sample1}}
  \subfigure[solution search]{
  \includegraphics[width=0.2\textwidth]{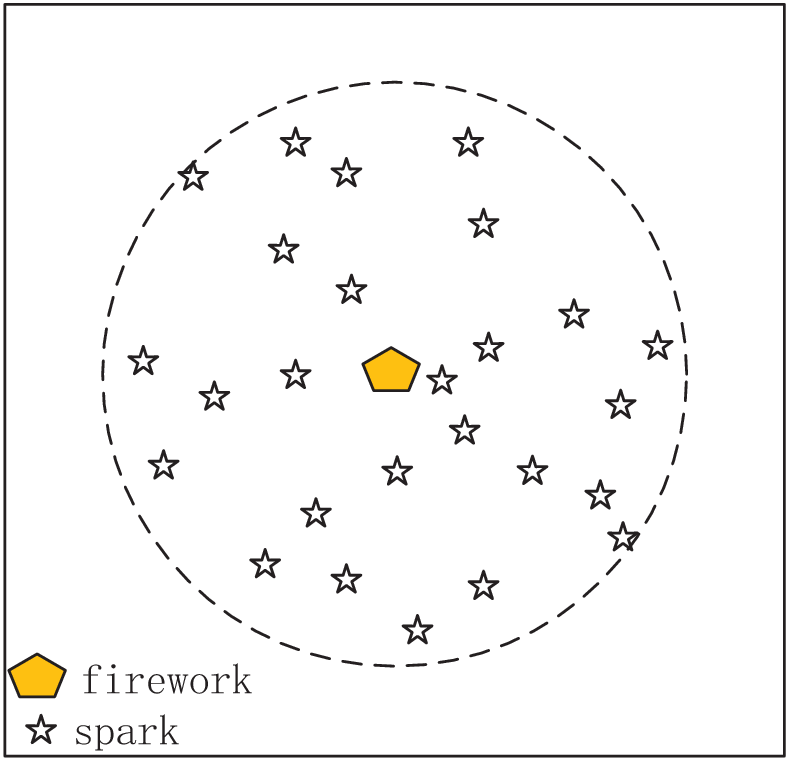}
  \label{Fig:sample2}}
  \caption{Comparison between fireworks explosion and solution search for optimization problems.}\label{Fig:sample}
\end{figure}

Inspired by the phenomenon of fireworks explosion in the night sky, a new SI algorithm called fireworks algorithm (FWA)~\cite{tan2010fireworks} has been proposed recently.
In FWA, each firework in the fireworks population performs a local search around its position, and cooperates with each other for the global search.
The explosion of fireworks can be seen as the search of solutions in the feasible range (see Fig. \ref{Fig:sample}). Since FWA has shown success in solving practical optimization problems, it is believed that FWA is efficient and worth for further study.

\subsection{Related work}
Related work based on FWA can be grouped into three categories, theory analysis,  algorithm developments and applications.
\subsubsection{Theory analysis}
In \cite{liu2014analysis}, Liu presented a theory analysis for FWA, and proved that FWA is an absorbing Markov stochastic process. Moreover, the convergence property and time complexity of FWA was also discussed.

\subsubsection{Algorithm developments}
Most studies focussed on algorithmic improvements of single objective FWA developments, multi-objective FWA developments and parallel FWA implementations.
%

%
In \cite{pei2012}, Pei \textit{et al.} investigated the influences of approximation landscape approaches on accelerating FWA by introducing the elite point of the approximation landscape into the fireworks swarm.
Additionally, Liu \textit{et al.}~\cite{liu2013improvement} investigated the influences on FWA when using new
methods for computing the explosion amplitude and the number of explosion sparks
while still maintaining the core idea that fireworks with smaller fitness will have more sparks and smaller explosion amplitudes.

The first comprehensive study of the operators of FWA can be found in~\cite{zhengCEC2013}, where Zheng S. \textit{et al.} proposed the enhanced fireworks algorithm (EFWA), which incorporates five modifications compared to conventional FWA: ($i$) a new minimal explosion amplitude check, ($ii$) a new operator for generating explosion sparks, ($iii$) a new mapping strategy for sparks which are out
of the search space, ($iv$) a new operator for generating Gaussian sparks, and ($v$) a new operator for selecting the population for the next iteration. The limitations in conventional FWA are presented and several improvements are proposed to deal with these limitations.

Based on the work of EFWA,
Zheng S.~\textit{et al.} proposed the dynamic search fireworks algorithm (dynFWA)~\cite{DynFWA}. In dynFWA, the firework with minimal fitness in each iteration is called \textit{core firework} and uses the dynamic explosion amplitude strategy while for the rest of fireworks, they use the same strategies as in EFWA (see \ref{ss:dynFWA} for the details of dynFWA). In addition,
Li \cite{AFWA} proposed an adaptive version of FWA (AFWA), in which the explosion amplitude of the firework with minimal fitness is calculated as the infinite norm distance between the firework and a certain candidate from the explosion sparks (see \ref{ss:afwa} for the details of AFWA).

Additionally, a number of hybrid algorithms between FWA and other algorithms have been proposed.
Gao \textit{et al.}~\cite{gao2011cultural} proposed cultural fireworks algorithm which is the hybrid between cultural algorithm and FWA.
Zheng Y. \textit{et al.}~\cite{zheng2012firework} proposed a hybrid algorithm FWA-DE of FWA and differential evolution (DE).
Zhang \textit{et al.}~\cite{6900289} proposed a hybrid biogeography-based
optimization with fireworks algorithm
while Yu \textit{et al.}~\cite{yuimprove} also presented a hybrid algorithm between DE with FWA.

For multi-objective FWA (MOFWA),
Zheng Y.\textit{et al.} proposed the framework of MOFWA which uses the fitness assignment strategy from \cite{zitzler2001spea2} and minimal pairwise distance metrics in\cite{hajek2010new}.

For Parallel FWA implementation, Ding \textit{et al.} proposed GPU-FWA~\cite{dingGECCO2013}, a GPU implementation which maintains a high speedup value. 

\subsubsection{Applications}
For the practical applications, FWA has been applied in FIR and IIR digital filters design~\cite{gao2011cultural}, the calculation of non-negative matrix factorization (NMF)~\cite{janecek2011using, janecek2011iterative, janecek2011swarm}, spam detection~\cite{he2013parameter}, image recognition~\cite{zhengICIST2013}, power system reconfiguration scheme~\cite{mohamed2014new}, mass minimisation of trusses~\cite{pholdee2014comparative}\cite{bureerat2011hybrid}, non-linear equation set~\cite{Duzhenxin2013} and 0/1 knapsack problems~\cite{Zhangjiaqin2011}.

All above applications indicate that FWA is one promising SI algorithm and has a profound value and wide future.

\subsection{Contributions}
The main contributions of this paper are:
($i$) an evaluation of the cooperative strategies among the fireworks of EFWA, dynFWA and AFWA is conducted.
($ii$) the proposal of the cooperative framework for FWA with independent selection method and crowdness-avoiding cooperative strategy.
In the cooperative framework, the independent selection method will ensure the information inheritance and improve the local search ability for each firework while the cooperative strategy can enhance the global search ability of the fireworks swarm.
($iii$) a comparison of the two different explosion amplitude strategies in dynFWA and AFWA, as well as the relationship between them are presented.

The remainder of this paper is organized as follows: Section~\ref{Sec:EFWA} introduces the framework of FWA and EFWA, and gives a comparison between FWA with other meta-heuristic algorithms.
Section~\ref{Sec:dynFWAvsAFWA} describes two state-of-the-art improvement works on FWA, \textit{i.e} dynFWA and AFWA, which focus on the explosion amplitude strategies.
In Section~\ref{Sec:newFramework}, we present a comprehensive analysis of cooperative strategies in conventional FWA framework. To overcome the limitations, the new cooperative framework of FWA is finally proposed.
To validate the ideas, several experiments are conducted in Section~\ref{Sec:ExperimentsDesign} and~\ref{Sec:ExperimentalResults}.
Finally, concluding remarks are given in Section~\ref{Sec:futurework}.

\section{The Framework of FWA and EFWA}\label{Sec:EFWA}

\subsection{General Framework}
A principal FWA works as follows: At first, $N$ fireworks are initialized randomly, and their quality (\textit{i.e.,} fitness) is evaluated in order to determine their \textit{explosion amplitude} (i.e., the explosion range) and the number of sparks for each firework. Subsequently, the fireworks explode and generate different types of sparks within their local space.
In order to ensure diversity and balance the global and local search, the explosion amplitude and the population of the newly generated explosion sparks differ among the fireworks. A firework with \textit{better} fitness can generate a \textit{larger population} of explosion sparks within a \textit{smaller range}, \textit{i.e.,} with a small explosion amplitude. Contrary, fireworks with lower fitness can only generate a smaller population within a larger range, \textit{i.e.,} with higher explosion amplitude. This technique allows to balance between exploration and exploitation capabilities of the algorithm. \textit{Exploration} refers to the ability of the algorithm to explore various regions of the search space in order to locate promising good solutions, while \textit{exploitation} refers to the ability to conduct a thorough search within a smaller area recognized as promising in order to find the optimal solution~\cite{clerc2002}. Exploration is achieved by fireworks with large explosion amplitudes (\textit{i.e.,} lower fitness), since they have the capability to escape from local minima. Exploitation is achieved by fireworks with small explosion amplitudes (i.e, high fitness), since they reinforce the local search ability in promising areas. After the explosion, another type of sparks are generated based on a Gaussian mutation of randomly selected fireworks. The idea behind this is to further ensure diversity of the swarm. In order to improve readability we assign new notations to the two distinct types of sparks: ``explosion sparks'' are generated by the explosion operator, and ``Gaussian sparks'' are generated by Gaussian mutation operator. To retain the information of the fireworks swarm, and pass it to the next iteration, a subset of the whole population is selected at the end of each iteration. The algorithm continues until the terminal criterion is met. The framework of (E)FWA is given in Alg.~\ref{Alg:EFWA}.

\begin{algorithm}[ht]
\caption{-- The general structure of (E)FWA.}
\label{Alg:EFWA}
\begin{algorithmic}[1]
	\STATE Initialize $N$ fireworks and set the constant parameters\;
	\REPEAT
	 \vspace{0.15cm}
    \STATE \textbf{// \textit{Explosion operator}}\
    \STATE Calculate the explosion amplitudes\;
	\STATE Calculate the numbers of explosion sparks\;
    \STATE Generate ``explosion sparks'' (check if out-of-bounds)
		\STATE Evaluate the fitness\\[0.5em]
%
		\STATE \textbf{// \textit{Gaussian mutation operator}}\
    \STATE Generate ``Gaussian sparks'' (check if out-of-bounds)\;
		\STATE Evaluate the fitness\\[0.5em]
%
    \STATE \textbf{// \textit{Selection strategy}}\
		\STATE  Select fireworks for the next iteration\\[0.5em]
	\UNTIL{termination criterion is met.}
\end{algorithmic}
\end{algorithm}

\subsection{Explosion Operator}
\label{ss:Explosion}
To perform the search for a firework, the points around the firework's position are sampled with uniform distribution ( taking the inspiration from fireworks explosion process in the nature, see Fig.~\ref{Fig:sample}). For this sampling operation, the sampling points number and the sampling range are two important factors which will influence the optimization results. The sampling points number is denoted as \textit{explosion sparks number} and the sampling range is denoted \textit{explosion amplitude}.

\subsubsection{Explosion amplitude and explosion sparks number}
\label{ss:AMP}
Assume that the optimization problem $f$ \footnote{\, In this paper, without loss of generality, the optimization problem $f$ is assumed to be a minimization problem.}
is a minimization problem, the fireworks number is $N$, the explosion amplitude $A$ (Eq.~\ref{Eq:Amplitude}) and the number of explosion sparks $s$ (Eq.~\ref{Eq:sonnum}) for each firework $X_i$ are calculated as follows.
\begin{equation}\label{Eq:Amplitude}
A_i=\hat{A} \cdot \frac{f\left(X_i\right)-y_{\mathrm{min}}+\varepsilon}{\sum_{i=1}^{N}\left(f\left(X_i\right)-y_{\mathrm{min}}\right)+\varepsilon}
\end{equation}
\begin{equation}\label{Eq:sonnum}
    s_i=M_e \cdot \frac{y_{\mathrm{max}}-f(X_i)+\varepsilon}{\sum_{i=1}^{N}(y_{\mathrm{max}}-f(X_i))+\varepsilon}
\end{equation}
where, $y_{\mathrm{max}}=\mathrm{max}(f(X_i))$, $y_{\mathrm{min}}=\mathrm{min}(f(X_i))$, and $\hat{A}$ and $M_e$ are two constants to control the explosion amplitude and the number of explosion sparks, respectively, and $\varepsilon$ is the machine epsilon. Additionally, the number of sparks $s_i$ that can be generated by each firework is limited by the lower/upper bounds.

Eq.~\ref{Eq:Amplitude} reveals that the explosion amplitude of the firework at the best location $X_b$ is usually very small [close to 0], the explosion sparks for $X_b$ will be located at the same location as $X_{b}$. As a result, these sparks cannot improve the location of $X_{b}$. To overcome this problem, EFWA used the \textit{minimal explosion amplitude check strategy} (MEACS) to bound the explosion amplitude $A_i$ of firework $X_i$ as follows.
\begin{equation}
A_i =
  \begin{cases}
   A_{\min} & \text{if } A_i < A_{\min},\\
   A_i      & otherwise,\\
  \end{cases}
\label{Eq:Amplitude_check}
\end{equation}
where, $A_{\min}^k$ decreases non-linearly with increasing number of function evaluations such that

\begin{equation}
	A_{\min}=A_{\mathrm{init}} - \frac{A_{\mathrm{init}}-A_{\mathrm{final}}}{E_{\mathrm{max}}}\sqrt{(2*E_{\mathrm{max}}-t\,)\,t},
	\label{Eq:nonlinear}
\end{equation}
where, $t$ refers to the number of function evaluation at the beginning of the current iteration, and $E_{\mathrm{max}}$ is the maximum number of evaluations. $A_{\mathrm{init}}$  and $A_{\mathrm{final}}$ are the initial and final minimum explosion amplitude, respectively.

\subsubsection{Generating explosion sparks}
\label{ss:EXP}
Now, each firework explodes and creates a different number of \textit{explosion sparks} within a given range of its current location. For each of the $s_i$ explosion sparks of each firework $X_i$, Algorithm~\ref{Alg:ExplosionSparkEFWA} is performed once.

\begin{algorithm}[ht]
\begin{algorithmic}[1]
  \STATE Initialize location of ``explosion sparks'': $\hat{x}_i = X_i$
  \STATE Set $z^k=\mathrm{round}(\mathrm{rand}(0,1))$, $k=1,2,...,d.$
  \FOR{\mbox{each dimension of} $\hat{x}_i$, where $z^k == 1$}
      \STATE Calculate offset displacement: $\triangle{X}^k=A_i \times \mathrm{rand}(-1,1)$
			\STATE $\hat{x}_{ik}=\hat{x}_{ik}+\triangle{X}^k$
			\IF{$\hat{x}_{ik}$ is out of bounds}
				\STATE randomly map $\hat{x}_{ik}$ to the potential space
			\ENDIF
	\ENDFOR
 \end{algorithmic}
 \caption{ -- Generating one ``explosion spark'' in EFWA}
 \label{Alg:ExplosionSparkEFWA}
 \end{algorithm}

\subsubsection{Mapping out-of-bounds sparks}
\label{ss:MAP}
When the location of a new explosion spark exceeds the search range in dimension $k$, this spark will be mapped to another location in the search space with uniform distribution (in dimension $k$) according to $\hat{x}_{ik}=X_{min}^k+\mathrm{rand}(0,1)*(X_{max}^k-X_{min}^k)$, here $\mathrm{rand}(a,b)$ denotes a random number from uniform distribution in $[a,b]$.

\subsection{Gaussian Mutation Operator}
\label{ss:gaussian}
In addition to the explosion sparks for local search, FWA uses another kind of sparks called \textit{Gaussian sparks}, to further ensure diversity of the swarm. Compared to the explosion sparks, Gaussian sparks have global search ability.

\begin{figure}[!h]
\centering
  \subfigure{
  \includegraphics[width=0.28\textwidth]{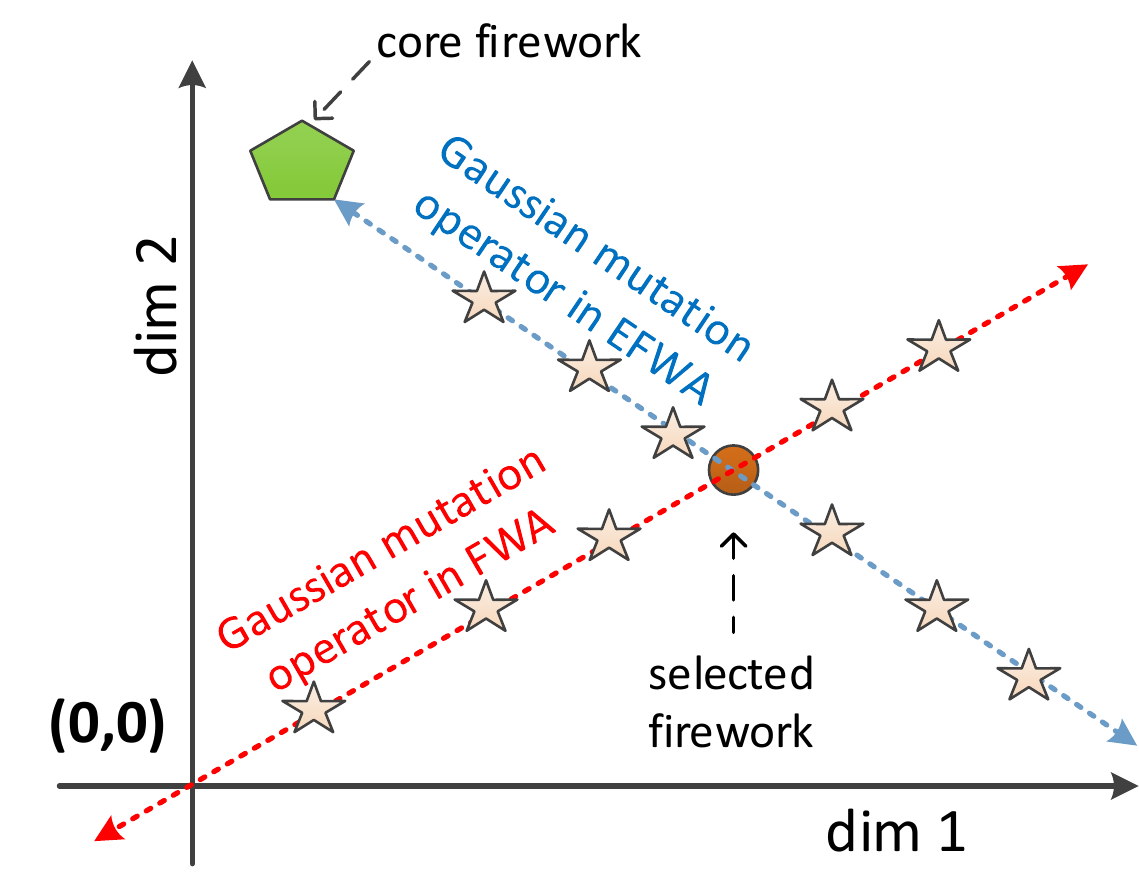}}
  \caption{Gaussian mutation operator in FWA and EFWA}
  \label{Fig:GaussianOperators}
\end{figure}

In the history of FWA, two typical Gaussian mutation operators were proposed. In FWA, the Gaussian mutation operator focuses on the mutation ability for each firework alone by multiplying a random value under Gaussian distribution with the positions of the firework in the selected dimensions. Later on in EFWA, the introduced Gaussian mutation operator emphasizes the cooperative evolution in the fireworks swarm (see Fig.~\ref{Fig:GaussianOperators}).

\subsubsection{Generating Gaussian sparks}
\label{ss:GAU}
Alg.~\ref{Alg:GaussianEFWA} gives some details about how Gaussian sparks are generated in EFWA. This algorithm is performed $M_g$ times, each time with a randomly selected firework $X_i$ ($M_g$ is a constant to control the number of Gaussian sparks, $Gaussian(a,b)$ denotes a random value from a normal distribution with expected value and variance set to $a$ and $b$, respectively).

\begin{algorithm}[ht]
 \begin{algorithmic}[1]
 \STATE Initialize the location of the ``Gaussian sparks'': $\bar{x}_i = X_i$
  \STATE Set $z^k=\mathrm{round}(\mathrm{rand}(0,1))$, $k=1,2,...,d$
   \STATE Calculate offset displacement: $e=\mathrm{Gaussian}(0,1)$
  \FOR{\mbox{each dimension} $\bar{x}_i$, where $z^k == 1$}
    \STATE $\bar{x}_{ik} = \bar{x}_{ik} + (X_{bk} - \bar{x}_{ik})*e$, where $X_{b}$ is the position of the best firework found so far.
  \IF{$\bar{x}_{ik} $ is out of bounds}
  \STATE $\bar{x}_{ik} =X_{min}^k+\mathrm{rand}(0,1)*(X_{max}^k-X_{min}^k)$
    \ENDIF
  \ENDFOR
 \end{algorithmic}
 \caption{ -- Generating one ``Gaussian spark'' in EFWA}
 \label{Alg:GaussianEFWA}
 \end{algorithm}

\subsubsection{Mapping out-of-bounds sparks}
When the location of a new Gaussian spark exceeds the search range in the $k$th dimension, the mapping method is similar as the explosion sparks which exceed the search range.

\subsection{Selection Strategy}
\label{ss:SEL}
To retain the information of the swarm and pass them to the next iteration, a subset of the whole population has to be selected.
EFWA applies a computationally efficient selection method called \textit{Elitism-Random Selection} (ERP,~\cite{engelbrecht2005}). The optimum of the set (\textit{i.e.} all fireworks, the explosion sparks, and the Gaussian sparks) will be selected firstly and the rest of individuals are selected randomly.
Although this method is rather simple compared to e.g., distance based selection method in FWA (cf.~\cite{lu2002improvement, tan2010fireworks}), our analysis in~\cite{zhengCEC2013} has shown that there is almost no difference in terms of convergence, final fitness and standard deviation between the distance based selection of conventional FWA and ERP.

\subsection{Comparison with Other Meta-heuristic Algorithms}
Since its introduction, FWA has shown great success for optimization problems, which indicates that some mechanisms of this algorithm are efficient. Here, we pick two famous algorithms, particle swarm optimization (PSO) and genetic algorithm (GA) for comparison.

\subsubsection{Reproduction Mechanism}
From the offspring generation view, we present the following definitions: ``\textit{cross-propagation}'' means that the offspring is generated by at least two parents, while ``\textit{self-propagation}'' means the offspring is generated by only one parent.

Thus, it can be seen that both cross-propagation and self-propagation are presented in GA, and there is no bijection from generation to generation.
In PSO, the position of the son particle is based on the information of this particle's history information and the global information of the best particle. Only the cross-propagation lies in PSO and there is a bijection from generation to generation.
The reproduction mechanism in FWA is similar to GA. The explosion sparks are generated by the explosion operator performed by one firework and mutation sparks are generated by the mutation operator, \textit{i.e.} both cross-propagation and self-propagation are presented in FWA and there is no bijection. However, different from the GA, the explosion operator in FWA plays the major role for optimization.

\subsubsection{Exploration and Exploitation Capacities}
Previous work in~\cite{eberhart1998comparison} has given the comparison between PSO and GA.
For GA, the crossover operator is usually very effective. At the early search phase, the crossover operator can move the chromosome to a very far space in the solution space, which makes GA maintain high exploration capacity.
While at the later phase, as the population of chromosomes converges, the crossover operator will not be able to move it to a far space as the two chromosomes usually have the similar structure, which makes the GA maintain high exploitation capacity. In addition, the mutation operator can move the chromosome to anywhere in the solution space which enhances the exploration capacity of GA.
In PSO, each particle is attracted by current position, its previous best position and the global best position (of the entire population). This has similar function as the crossover operator in GA~\cite{eberhart1998comparison}, and finally the whole population of particles will converge to one position, which makes the swarm maintain a high exploitation capacity. The particles have to move one by one step from one place to another, which cannot jump with a quite high step to another place in one iteration~\cite{eberhart1998comparison}. However, it seems that GA and FWA can do this by uding a big mutation rate and a big explosion amplitude, respectively.
For FWA, the exploration and exploitation capacities are gained by taking different explosion amplitudes and explosion sparks number for each firework. A small explosion amplitude with large explosion sparks number can make the firework perform a comprehensive local search.
To some extend, FWA is quite similar to GA, the smaller explosion amplitude has the
similar effect as mutation operator with small mutation rate while
the big amplitude has the similar effect as mutation operator with big mutation rate.

\subsubsection{Survival strategy}
In GA, the selection strategy is performed in each iteration to maintain and pass the information to the next iteration.  The fittest candidate will be surely selected to the next iteration, while for the rest of candidates, they are selected with probability, which means that there is no bijection from generation to generation.
For PSO, it does not utilize the selection operator, and there is bijection from generation to generation for each particle.
In FWA, the selection operator is performed like GA in each iteration. Section \ref{S:convential_coo} will give a detailed discussion with the survival strategy.

\section{State-of-the-art Improvement Work: dynFWA and AFWA}
\label{Sec:dynFWAvsAFWA}

\subsection{Definition}
For the convenience of  discussion, the following definitions are introduced at first:

\medskip
\noindent
\emph{Core firework (CF) and non-core firework (non-CF)}:  Among the fireworks swarm, the firework with minimal fitness according to the $f$ is denoted as CF. All other fireworks except the CF are denoted as non-CFs.

\medskip
\noindent
\emph{General core firework (GCF) and non-general core firework (non-GCF)}:
The fireworks which have the same parent firework as current CF may have the similar performance as the CF due to the selection method. Thus, we extend the CF to general CF by including the fireworks which have the same parent as CF. The detailed definition is given below.

In each iteration, after the generation of explosion sparks, in the candidates set ($S_c$), which includes the fireworks and explosion sparks, the selection operation is performed.
Let us define $x_b$ as the ``best'' newly created explosion spark of all fireworks in the swarm, and $X_{CF}$ as the current core firework.
Assume that $P(x_i)$ denotes the firework which generated $x_i$, and $S_{GCFs}$ denotes the set of GCF.
\begin{itemize}
  \item If $f(x_b)-f(X_{CF})<0$, and the candidate $x_i$ which is from the candidate set $S_c$ is subjected to the following condition,
  \[x_i = \mathop{\mathrm{arg}}\limits_{x_i} \quad (P(x_i)==P(x_b)) || (x_i == P(x_b))\]
  and $x_i$ is selected as firework into next iteration, then $x_i \in S_{GCFs}$.
  \item If $f(x_b)-f(X_{CF}) \geq 0$, and the candidate $x_i$ which is from the candidate set $S_c$ is subjected to the following condition,
  \[x_i = \mathop{\mathrm{arg}}\limits_{x_i} \quad (P(x_i)== X_{CF}) || (x_i \in S_{GCFs})\]
  and $x_i$ is selected as firework into next iteration, then $x_i \in S_{GCFs}$.
\end{itemize}
For the non-GCF, among the fireworks swarm, the fireworks except for GCF are denoted as non-GCFs.

\subsection{Explosion Strategies in dynFWA and AFWA}
In EFWA, the MEACS (ref Section~\ref{ss:Explosion}) enforces the exploration capabilities at the early phase of the algorithm (larger $A_{\min}$ $\Rightarrow$ global search), while at the final phase of the algorithm the exploitation capabilities are enforced (smaller $A_{\min}$ $\Rightarrow$ local search). Additionally, the non-linear decrease of $A_{\min}$ (cf. Eq.~\ref{Eq:nonlinear}) enhances exploitation already at an earlier stage of the EFWA algorithm. However, this procedure decreases the explosion amplitude solely with the current number of function evaluations which heavily depends on the pre-defined number of iterations for the algorithm. 
The explosion amplitude strategy should consider the optimization process information and the information of the explosion sparks rather than solely the information about the current iteration [evaluation] count. To approach this problem, the adaptive explosion amplitude strategies for fireworks were proposed, which are \textit{dynamic search in FWA} (dynFWA) and \textit{adaptive FWA} (AFWA).

In dynFWA and AFWA, the fireworks are separated into two groups. The first group consists of the CF, while the second group consists
of all remaining fireworks. The responsibility of the CF is to perform a local search around the best location found so far,
while the responsibility of the second group is to maintain the global search ability.

\begin{algorithm}[ht]
 \begin{algorithmic}[1]
 \REQUIRE Define:\\
	${X}_{CF}$ is the current location of the CF;\\
	$x_{b}$ is the best location among all explosion sparks;\\
	$A_{CF}^t$ is the CF's explosion amplitude in iteration $t$;	\\
	$C_a$ is the amplification factor;	\\
	$C_r$ is the reduction factor;	\\
 \ENSURE
  \IF{ $f({x}_{b}) - f(X_{CF}) <0$}
  \STATE $A_{CF}^t \leftarrow A_{CF}^{t-1} \times C_a$;
  \ELSE
  \STATE $A_{CF}^t  \leftarrow A_{CF}^{t-1} \times C_r$;
  \ENDIF
 \end{algorithmic}
 \caption{-- dynFWA explosion amplitude update strategy}
 \label{Alg:dynFWA}
 \end{algorithm}

\subsubsection{The dynFWA}
\label{ss:dynFWA}

Let us define ${x}_{b}$ as the ``best'' newly created explosion spark of all fireworks in the swarm, for the dynFWA, the explosion amplitude adapts
itself according to the quality of the generated sparks.

\smallskip
\noindent
\textbf{Case 1)} : One or several explosion sparks have found a better position.
It is possible that ($i$) an explosion spark generated by the CF has found the best position, or that  ($ii$) an explosion spark generated by a \textit{different} firework than the CF has found the best position. Both cases indicate that the swarm has found a new promising position and that ${x}_{b}$ will be the CF for the next iteration.
	
($i$) In most cases, $x_{b}$ has been created by the CF. In such cases, in order to speed up the convergence of the algorithm, the explosion amplitude of the CF for the next iteration will be increased compared to the current iteration.
	
($ii$)  In other cases, a firework different from the CF will create ${x}_{b}$. In such cases, ${x}_{b}$ will become the new CF for the next iteration. Since the position of the CF is changed, the current explosion amplitude of the current CF will not be effective to the newly selected CF.
However, it is possible that ${x}_{b}$ is located in rather close proximity to the previous CF due to the fact that the random selection method may select several sparks created by the CF, which are initially located in close proximity to the CF. If so, the same consideration as in ($i$) applies. If ${x}_{b}$ is created by a firework which is not in close proximity to the CF, the explosion amplitude can be re-initialized to the pre-defined value. However, since it is difficult to define ``close'' proximity, we do not compute the distance between ${x}_{b}$ and $X_{CF}$ but rely on the dynamic explosion amplitude update ability. Similarly to ($i$), the explosion amplitude is increased. If the new CF cannot improve its location in the next iteration, the new CF is able to adjust the explosion amplitude itself dynamically.

We underline the idea why an increasing explosion amplitude may accelerate the convergence speed: Assume that the current position of the CF is far away from the global/local minimum.
Increasing the explosion amplitude is a direct and effective approach in order to increase the step-size towards the global/local optimum in each iteration, \textit{i.e.}, it allows for faster movements towards the optimum. However, we should make sure that the fireworks walk towards to global/local optimal position. In fact, the probability to find a position with better fitness decreases with increasing explosion amplitude due to the increased search space, it will be a hard job and a challenge for fireworks.

\smallskip
\noindent
\textbf{Case 2)} : None of the explosion sparks of the CF nor of all other fireworks has found a position with better fitness compared to the CF, \textit{i.e.}, $\Delta_f >= 0$.
Under this circumstance, the explosion amplitude of the CF is reduced in order to narrow down the search to a smaller region around the current location and to enhance the exploitation capability of the CF. The probability for finding a position with better fitness usually increases with the decreasing of explosion amplitude.

\subsubsection{The AFWA}
\label{ss:afwa}
In AFWA, the motivation is to calculate the most suitable amplitude for the CF without preseted parameters by making full use of the region's information based on the generated explosion sparks around the current position~\cite{AFWA}.

 AFWA aims at finding a specific spark and using its distance to the best individual (which is either the $X_{CF}$ or $x_b$, and will be selected as the CF for the next iteration) as the explosion amplitude for the next round. The specific spark that AFWA chooses should be subject to the following two conditions:

 \begin{itemize}
    \item Its fitness is worse than the one of the current CF, which guarantees that the specific spark is not too close to CF;
    \item The distance between the specific spark with the best individual is minimal among all the candidates, which is to ensure the convergence.
 \end{itemize}

As in FWA, the explosion sparks are generated independently in each dimension, the \textit{infinite norm} is taken as the distance measure as follows.
\begin{equation}
  \left\| {\bf{x}} \right\|_\infty {\rm{ = }}\mathop {\max }\limits_i (\left| {{x_i}} \right|)
\end{equation}

The AFWA's explosion amplitude update strategy includes two cases:

\smallskip
\noindent
\textbf{Case 1)}
The CF does not generate sparks with smaller fitness than the firework's, \textit{i.e.} the \textit{best fitness of all the explosion sparks} $x_{CF,b}$ is worse than the core firework's fitness $X_{CF}$, and the explosion amplitude will be set to the infinite norm between the core firework and the specific spark. If so, the explosion amplitude in the next iteration will be reduced according to Alg.~\ref{Alg:AFWA}.

\smallskip
\noindent
\textbf{Case 2)}  The generated $x_{CF,b}$ has smaller fitness than $X_{CF}$, 
according to Alg.~\ref{Alg:AFWA}, it will be hard to determinate whether the explosion amplitude will be increased or decreased.
Generally, the explosion amplitude will be increased with high probability while be reduced with smaller probability.

Moreover, as the infinite norm between the specific spark and the firework changes radically, AFWA introduces the smoothing strategy ($A_{CF}^t \leftarrow 0.5 \cdot (A_{CF}^{t-1}+\lambda \cdot A_{CF}^t)$).

\begin{algorithm}
\caption{AFWA explosion amplitude update strategy}
\label{Alg:AFWA}
\begin{algorithmic}[1]
\REQUIRE Define:\\
	$M$ is explosion sparks number of CF;\\
	$A_{CF}^t$ is the CF's explosion amplitude in iteration $t$;\\
    $x_{CF,k}$ is the $k$-th explosion spark of CF;\\
    $x_{CF,b}$ is the best explosion spark of CF;\\
    $\lambda$ is a constant parameter which controls the update rate.
 \ENSURE
\FOR{$k=1$ \TO $M$}
    \IF {$||{{x}_{CF,k}}-{{x}_{CF,b}}|{|_\infty }>A_{CF}^t$ \AND $f({{x}_{CF,k}})>f({X_{CF}})$}
        \STATE $A_{CF}^t \leftarrow ||{{x}_{CF,k}}-{{x}_{CF,b}}|{|_\infty}$
    \ENDIF
\ENDFOR
\STATE $A_{CF}^t \leftarrow 0.5 \cdot (A_{CF}^{t-1}+\lambda \cdot A_{CF}^t)$
\end{algorithmic}
\end{algorithm}

\subsection{dynFWA vs AFWA}

\begin{figure}[t]
\centering
  \subfigure[dynFWA, the explosion amplitude is updated when no better solutions are found.]{
  \includegraphics[width=0.225\textwidth]{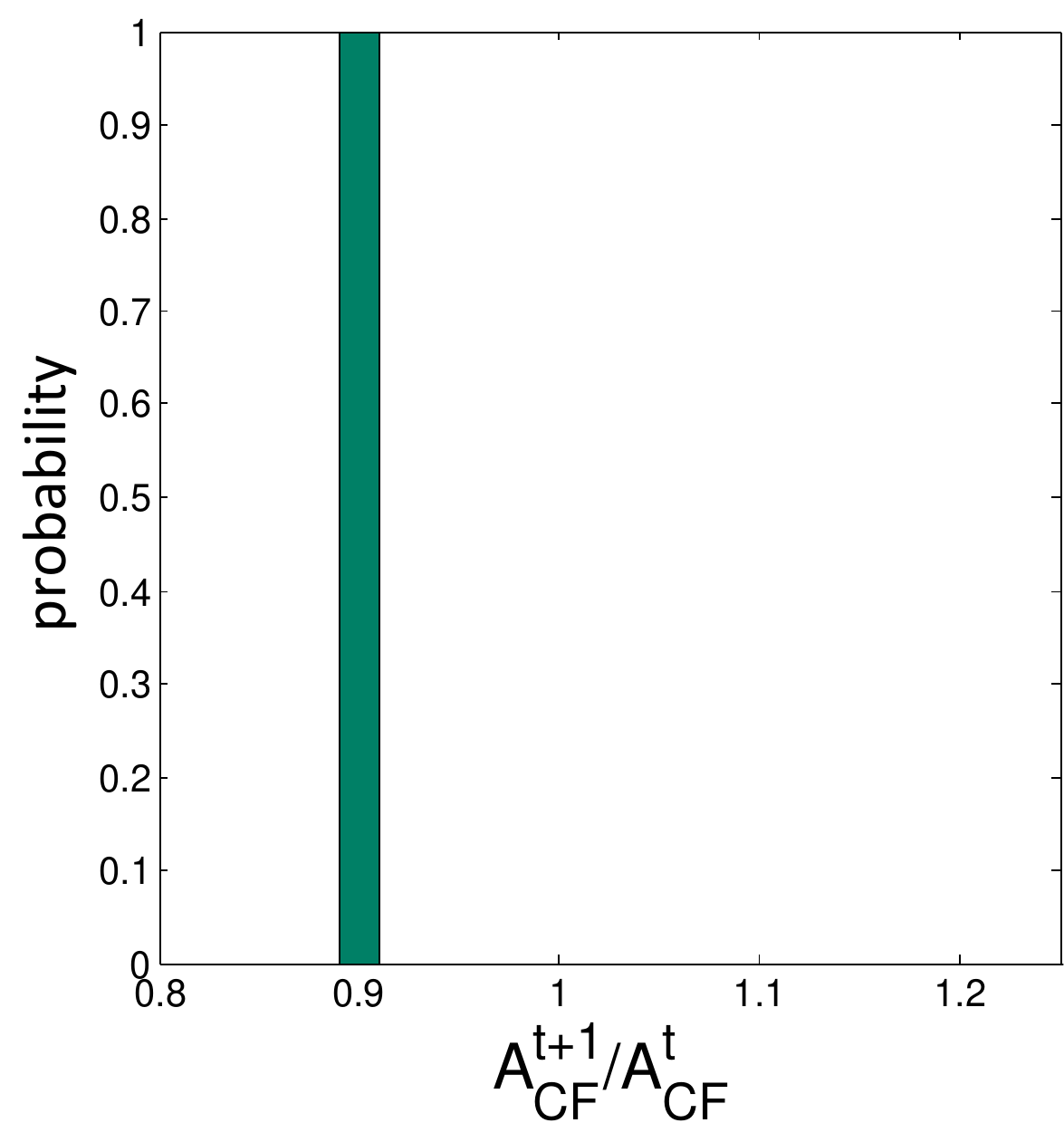}
  \label{Fig:dynFWA_Reduction}}
    \subfigure[dynFWA, the explosion amplitude is updated when a better solution is found.]{
  \includegraphics[width=0.225\textwidth]{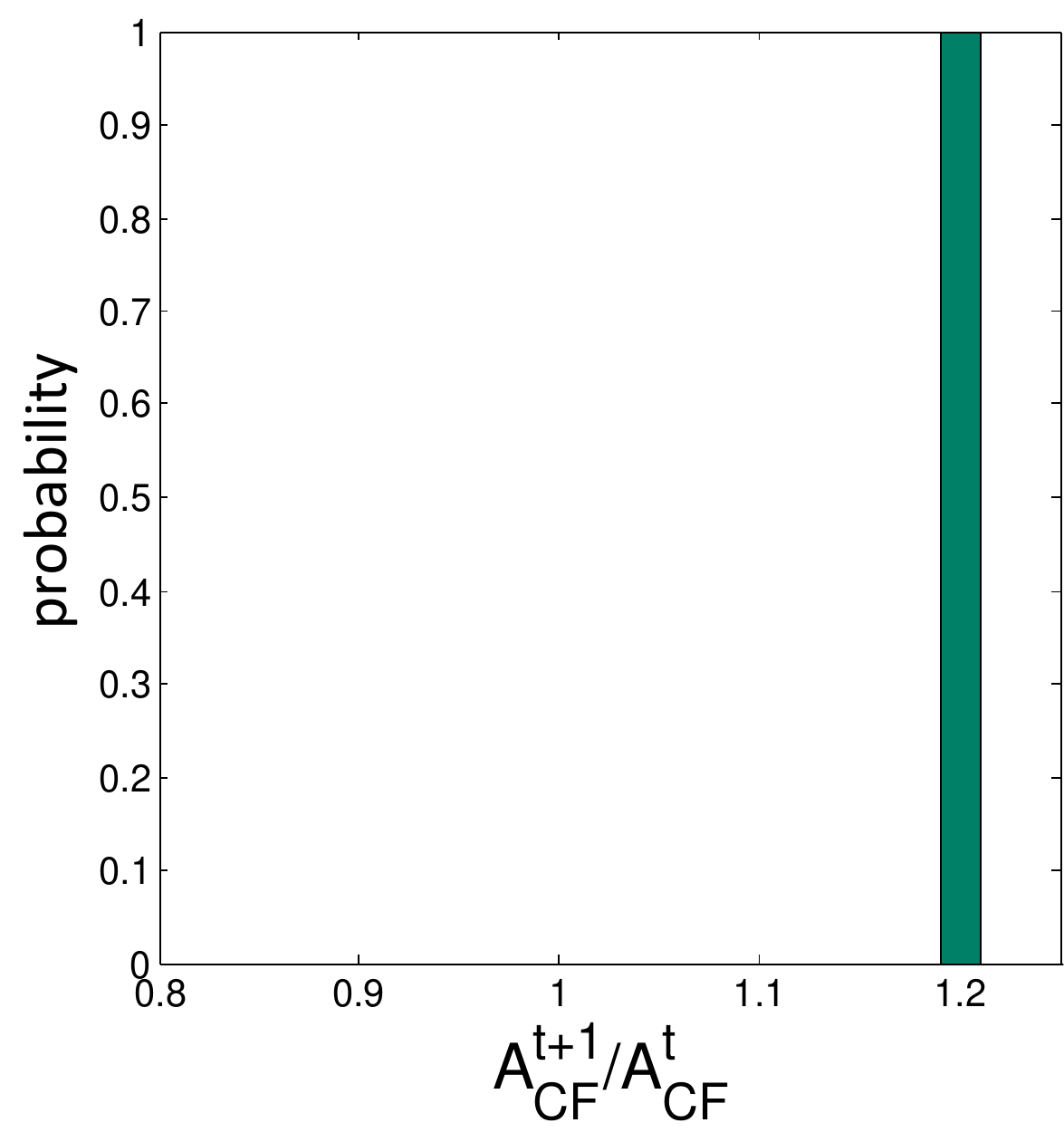}
  \label{Fig:dynFWA_Amplification}}\\
  \subfigure[AFWA, the explosion amplitude is updated when no better solutions are found.]{
  \includegraphics[width=0.225\textwidth]{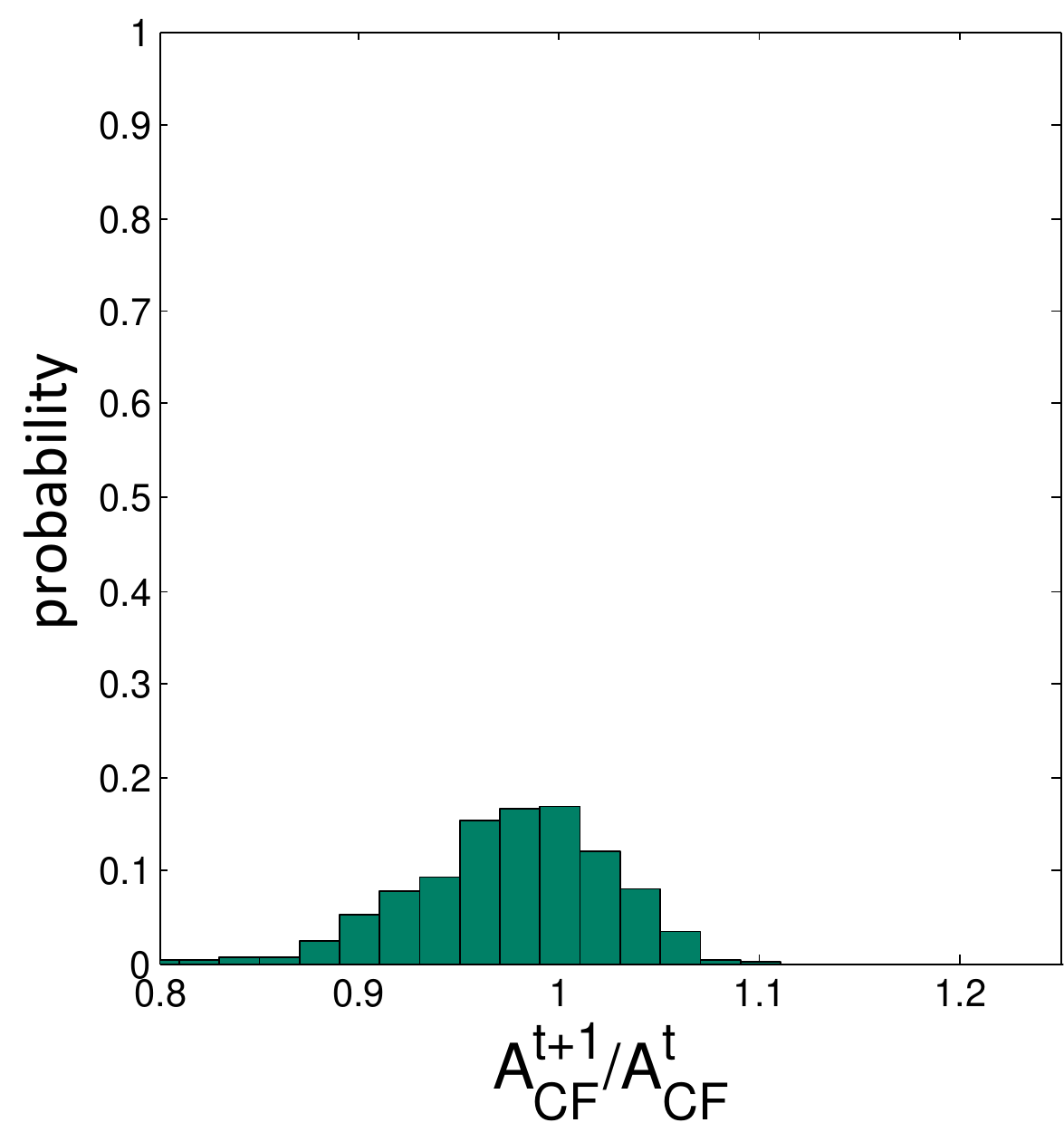}
  \label{Fig:AFWA_Reduction}}
    \subfigure[AFWA, the explosion amplitude is updated when a better solution is found.]{
  \includegraphics[width=0.225\textwidth]{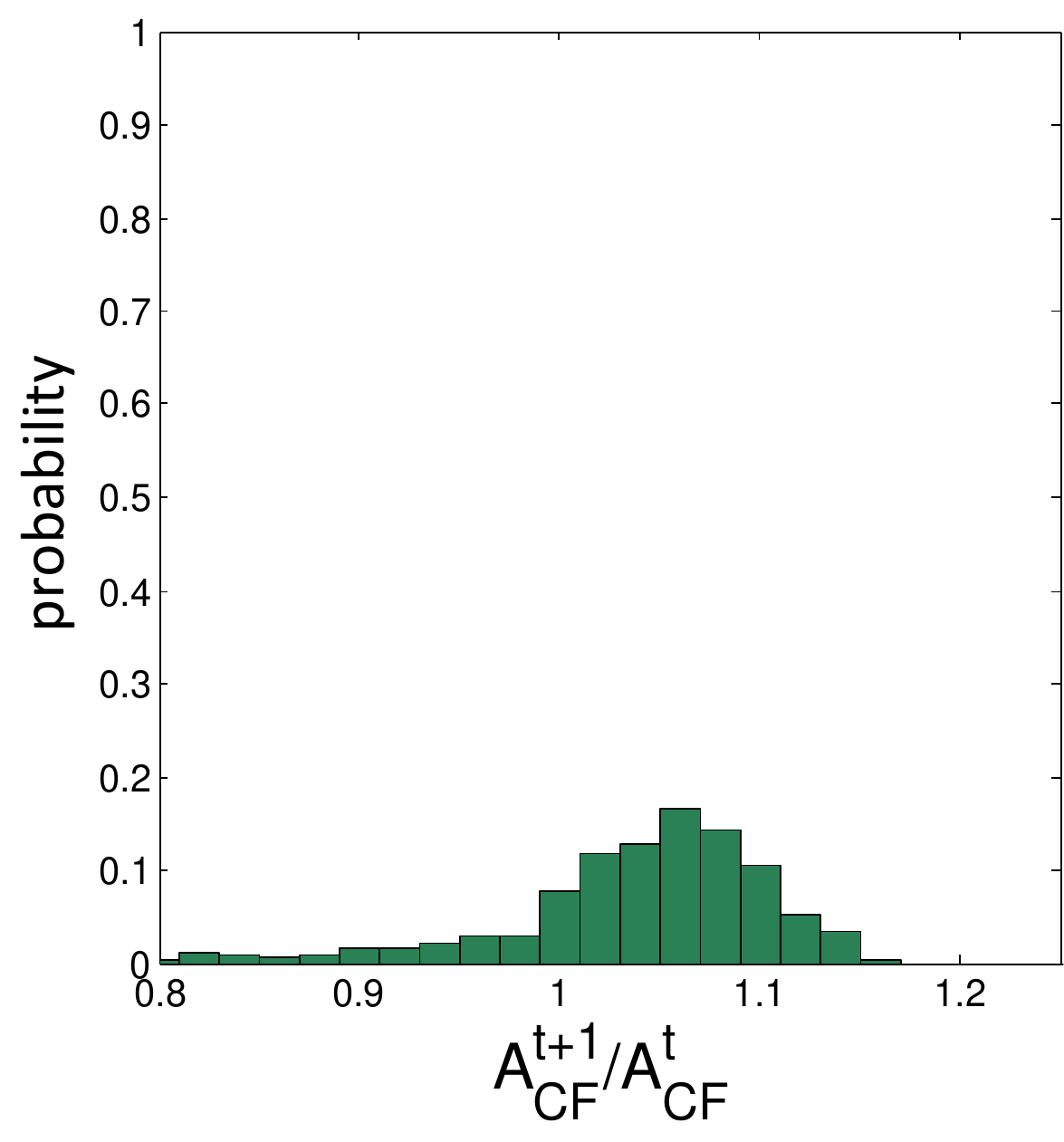}
  \label{Fig:AFWA_Amplification}}
  \caption{The comparison of explosion amplitude update between AFWA and dynFWA.}\label{Fig:ExplosionUpdate}
\end{figure}

From Alg.~\ref{Alg:dynFWA} and Alg.~\ref{Alg:AFWA}, it can be seen that if the CF cannot generate any sparks with better fitness, then the explosion amplitude of the firework will
be reduced in both AFWA and dynFWA.
If the CF has generated a spark with better fitness, the explosion amplitude of the firework in dynFWA will be amplified.
However, for AFWA, the determination of amplification or reduction is more complex.

To investigate the relationship between of the amplification and reduction with whether a better solution is found or not,
we record the values of $A_{CF}^{t}/A_{CF}^{t-1}$.
For dynFWA, if a better solution is found, the recorded value $A_{CF}^{t}/A_{CF}^{t-1}$ will be $C_a$ and the explosion amplitude is updated with the amplification factor.
Otherwise, the recorded value $A_{CF}^{t}/A_{CF}^{t-1}$ will be $C_r$ and the explosion amplitude is updated with the reduction factor (cf. Fig.~\ref{Fig:dynFWA_Amplification} and Fig.~\ref{Fig:dynFWA_Reduction}).

To investigate the explosion amplitude update process of AFWA, Sphere function is chosen as the benchmark function to study the
explosion amplitude amplification and reduction when the firework has found or not found a better solution.

Fig.~\ref{Fig:AFWA_Amplification} presents the histogram of $A_{CF}^{t}/A_{CF}^{t-1}$ in the iterations when a better solution is found, here the y-coordinate denotes the probability.
Fig.~\ref{Fig:AFWA_Reduction} presents the histogram of $A_{CF}^{t}/A_{CF}^{t-1}$ in the iterations when no better solutions are found.
The geometric mean in Fig.~\ref{Fig:AFWA_Amplification} is 1.017, while the geometric mean in Fig.~\ref{Fig:AFWA_Reduction} is 0.976.
It can been seen that if the firework has found a better position with smaller fitness,
the explosion amplitude in AFWA will be increased with higher probability, and reduced with smaller probability.

\begin{figure}[t]
  \centering
  \includegraphics[width=0.5\textwidth]{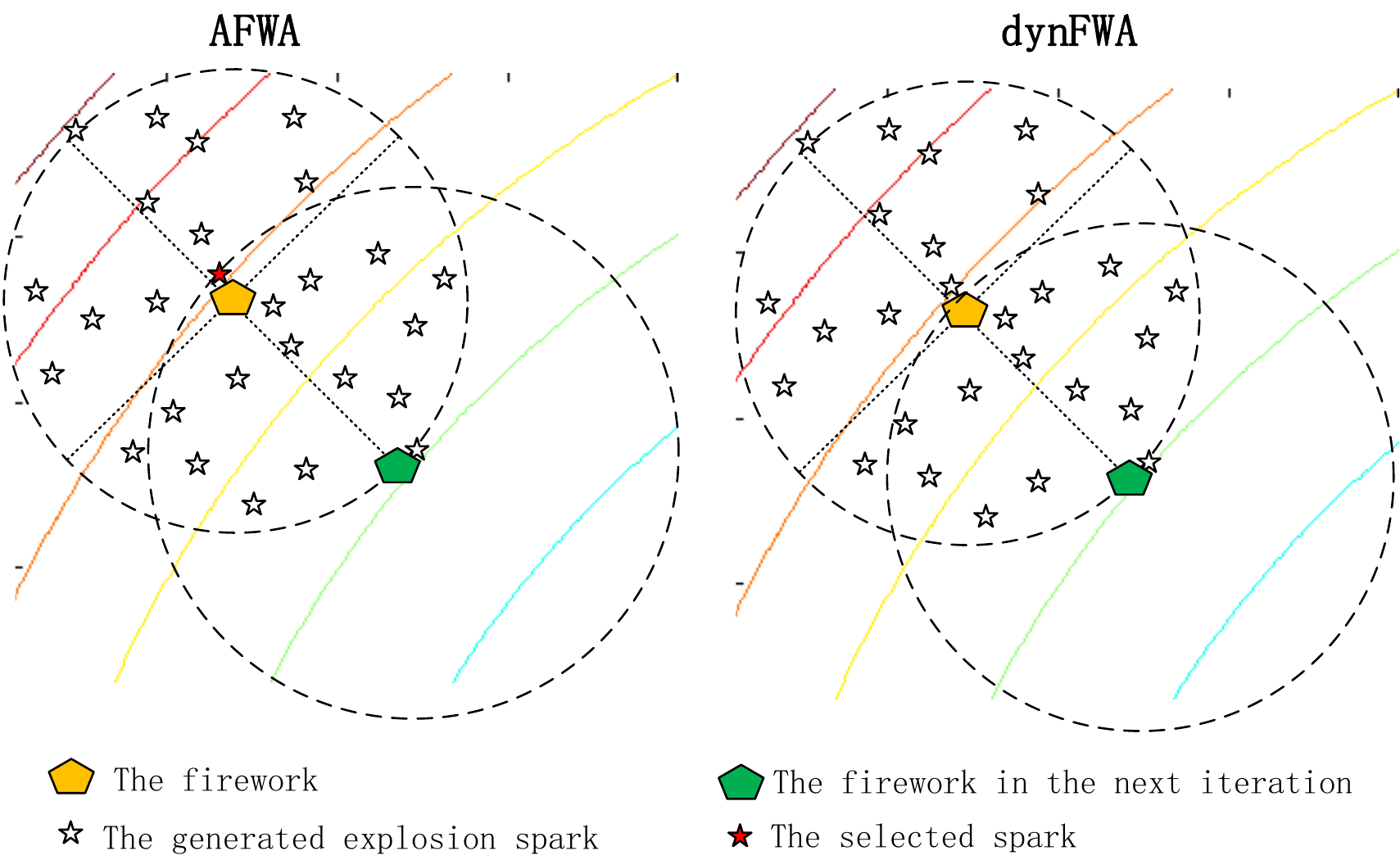}
  \caption{dynFWA vs AFWA in a special case}\label{Fig:special}
\end{figure}

Here, a special case is also considered to compare the two explosion amplitude strategies.
Assume that explosion sparks number $M$ is a very big value, \textit{i.e.} $M\rightarrow +\infty$, and the generated sparks are located in a float and continues region. Under this circumstance,
if the generated sparks have found any better positions, usually the best explosion spark will be located at the edge of the region (see Fig.~\ref{Fig:special} with dimension is set to 2),
then both AFWA and dynFWA will increase the explosion amplitude for the next iteration according to Alg.\ref{Alg:dynFWA} and Alg.\ref{Alg:AFWA}.

Thus, we can draw the following conclusions:
\begin{itemize}
  \item The explosion amplitude is one of the most crucial parameters for FWA to balance the global and local search ability.
  \item From the statistical view, the two methods -- dynFWA and AFWA are from different start points to adjust the explosion amplitude, but in a sense, finally they reach the same goal by different means.
\end{itemize}

\section{The Proposed FWA Cooperative Framework}
\label{Sec:newFramework}
In this section, we first give a detailed analysis of the cooperative strategies in conventional FWA framework and the limitations are pointed out, and then the new cooperative FWA framework (CoFFWA) is finally proposed, which includes the independent selection method for each firework and crowdness-avoiding cooperative strategy among the fireworks.

\subsection{Analysis of Cooperative Strategies in Conventional FWA Framework}
\label{S:convential_coo}
FWA is designed as one swarm intelligence algorithm, in which the fireworks in the swarm cooperate with each other to deal with the task that one firework cannot work well on.
In EFWA, two cooperative operations are used for the implementation of this idea:
\begin{itemize}
    \item the sharing of the fitness values in the population for the calculation of explosion amplitudes and explosion sparks number in the explosion operator (cf subsection \ref{ss:Explosion});
    \item the Gaussian mutation strategy in the Gaussian mutation operator (cf subsection \ref{ss:gaussian}).
\end{itemize}
The sharing of fitness values among the fireworks to calculate the explosion amplitudes and sparks numbers makes the fireworks with smaller fitness have smaller explosion amplitudes and larger number of explosion sparks, which maintains the capacity of exploitation while the fireworks with bigger fitness have bigger explosion amplitudes and smaller numbers of explosion sparks, which maintains the capacity of exploration. As for the Gaussian mutation operator, the generated Gaussian mutation sparks will locate along the direction between the core firework and the selected firework which is assumed to improve the diversity of the fireworks swarm.

However, for EFWA, dynFWA, and AFWA, all of them take the selection method from the \textit{candidates} set  which includes the fireworks in current iteration, the generated explosion sparks and Gaussian mutation sparks in each iteration.
For all of the FWA variants, the basic principle of the selection method is that the optimum among the \textit{candidates} is always kept while for the rest of fireworks, different methods have different selection probability calculation methods.

In the following, we will present an observation that these kinds of selection methods result in the fact that in the explosion operator the non-CFs contribute less for the optimization while taking up the most evaluation times,
and the Gaussian mutation operator proposed in EFWA is not as effective as it was designed to be.

\subsubsection{Cooperative Performance Analysis of CF and non-CFs in Explosion Operator}
In FWA, the calculation method of explosion amplitude will make the explosion amplitude of the CF close to zero. To avoid this limitation, EFWA introduces the MEACS (cf Section ~\ref{ss:Explosion}), where the CF's explosion amplitude is in fact calculated based on the MEACS which is not relevant to the non-CFs in the population.
In dynFWA or AFWA, the situation is similar, that the explosion amplitude of CF is calculated based on the dynamic search strategy or adaptive strategy which is also
not relevant to the non-CF's fitness.
Compare the CF in EFWA, dynFWA and AFWA, it can be seen that the CF's explosion amplitude strategies are independent of the non-CFs' in the fireworks swarm, the only interaction between the CF and non-CFs except for the selection method is the calculation of explosion sparks number which is powerless for dealing with complex problems.

Comparing the CF with non-CFs, the differences between them lie in two aspects, one is that the explosion amplitudes' calculation methods, the other is the selection method. For the explosion amplitude calculation method, different explosion amplitude calculation methods are taken to maintain the diversity of the fireworks swarm. However, the selection method makes the initial idea that the non-CFs are to increase the diversity of the fireworks swarm not effective.

In the selection, most of the sparks generated by CF and non-CFs are brought together. The \textit{candidate with minimal fitness} [will be the CF in the next iteration] is always selected at first, while for the rest of fireworks [will be non-CFs in the next iteration], they are selected with probability.
If a selected non-CF candidate is generated by the CF, then it will have similar performance as the CF.
Alternatively, the selected candidate is generated by the non-CFs.
In this case, the non-CFs are usually generated by the parent fireworks
within large explosion explosion amplitudes which can be seen as randomly generating of sparks within the search
range as the information of the non-CF can only be passed to the next iteration within few iterations with the conventional selection method, thus after a number of iterations, the position of the non-CF will be reinitialized in a new position and will be kept only in  few iterations.

In fact, these kind of selection strategies will make the generated sparks by the non-CFs have the similar
performance with randomly generated sparks in the search space to some extend or have the similar performance with sparks generated by CF.
We cannot expect the good fitness by randomly
generating the sparks if there is no much heuristic information.

In Section~\ref{Sec:ExperimentsDesign}, experiments are designed to evaluate the performance of CF and GCF to validate our idea that the non-CFs and non-GCFs make little contribution to the optimization.

\subsubsection{Cooperative Strategy in Gaussian Mutation Operator}
The motivation of Gaussian sparks is to increase the diversity of the swarm. The prerequisites for successful performance by cooperating in the fireworks population is that the information of each firework is different and effective.

In EFWA, the newly generated sparks locate along the direction between the selected firework ($X_i$) and CF ($X_{CF}$).
For the selected firework $X_i$, it can be classified into two categories. The first category comprises fireworks which are very close to $X_{CF}$, usually these fireworks have the same parent firework as the CF. If so, the newly generated Gaussian sparks may have similar performance with the explosion sparks which cannot increase the diversity of the swarm effectively. The other category is the fireworks which are not close to the CF, usually it is because these fireworks are generated by a firework different from the parent firework of CF.
If so, the newly generated Gaussian sparks will be $(i)$ close to CF $(ii)$ close to the selected firework, $(iii)$ not close to any
fireworks. If the newly generated Gaussian spark is close to either of the CF or the selected firework, it has similar
performance with the explosion sparks generated by them. If the generated Gaussian spark is not close to them, then it
can be seen as the spark generated by firework with large explosion amplitude. Thus, based on above analysis, the
newly generated Gaussian spark will not work effectively and is not be able to increase the diversity of the fireworks population.

Moreover, assume that the selected firework $X_i$ is a non-CF with a random position due to the selection method, then the generated Gaussian mutation spark will have the similar effects with randomly generating a spark in the search range under the case $(ii)$ and case $(iii)$. Thus, the generated Gaussian sparks will not be able to improve the diversity of the fireworks swarm.

\subsubsection{Conventional FWA Framework vs Evolutionary Strategy}
In the previous sections, we have pointed out that the non-CFs and Gaussian sparks do not contribute much to the optimization due to the selection method in the conventional FWA framework. If we simply eliminate the Gaussian mutation operator and non-CFs in the explosion operator, we will get the \textit{minimalist fireworks algorithm} (MFWA) using only one firework, as shown in Alg. \ref{Alg:MFWA}.
\begin{algorithm}
\caption{The MFWA}
\label{Alg:MFWA}
\begin{algorithmic}[1]
\STATE Initialize the firework and explosion amplitude
\REPEAT
\STATE Generate explosion sparks (cf. Alg.~\ref{Alg:ExplosionSparkEFWA})
\STATE Update explosion amplitude (cf. Alg.~\ref{Alg:dynFWA} or Alg.~\ref{Alg:AFWA})
\STATE Select the best individual as the new firework
\UNTIL{termination criterion (time, max. \# evals, convergence, ... ) is met}
\end{algorithmic}
\end{algorithm}

 For comparison, we briefly describe a kind of evolutionary algorithms: \textit{evolutionary strategy} (ES)~\cite{beyer2002evolution}. Generally speaking, ES conducts the search process by keeping the iterations of sampling and learning. In the sampling step, a number of points are randomly sampled from a Gaussian distribution. In the learning step, the parameters of the Gaussian distribution are updated according to the qualities of the sampled points.

The iteration process of MFWA is very similar to a $(1+\lambda)$-ES: one parent generates many sons, and the sampling parameters are updated according to the qualities of them (the position and the fitness). The explosion amplitude in FWA and the variance in ES can both be regarded as the step size. The methodology to control them are also very similar: dynamic explosion amplitude is similar to the 1/5 rule in $(1+1)$-ES, while adaptive explosion amplitude is comparable to $(1+\lambda)$-ES. So the properties of these algorithms are similar: they are all locally convergent, they all work poor in dimension sensitive functions, and they are all easily trapped in local minima.
After the comparison with ES, it leaves a profound insight that the new FWA framework which can utilize all the fireworks information should be designed and
the powerful and efficient cooperative mechanisms are essential for the future developments.

\subsection{Cooperative Framework of FWA}
To make FWA be a successful swarm intelligence algorithm~\cite{engelbrecht2005}, the information for each firework should be passed to the next iteration and fireworks can cooperate with each other for the optimization.

However, the conventional FWA framework lacks the local search ability for non-CFs, while the cooperative strategy in the Gaussian mutation operator is not very effective. To tackle these limitations, the CoFFWA with independent selection method and cooperative strategy is finally proposed.

\begin{algorithm}[ht]
\caption{-- The Cooperative Framework for FWA}
\label{Alg:NEWFramework}
\begin{algorithmic}[1]
	\STATE Initialize $N$ fireworks
	\REPEAT
    \FOR{each firework}
	   \STATE Generate the explosion sparks
       \STATE Select best candidate among its generated sparks and itself
       \STATE Perform the cooperative strategy
    \ENDFOR
	\UNTIL{termination criterion (time, max. \# evals, convergence, ... ) is met}
\end{algorithmic}
\end{algorithm}

\subsubsection{Independent Selection Method}
The analyses in previous subsection suggest that the previous random selection methods in FWA and its variants cause that non-CFs and non-GCFs do not contribute much to the optimization for a problem while consuming a lot of resources, \textit{i.e.} evaluation times in the explosion operator, and the cooperative scheme of Gaussian mutation operator among the fireworks is powerless to solve complex problems.
Furthermore, the selection method in conventional FWA framework is seen as the main reason why the non-CFs/non-GCF cannot evolve for a number of consecutive iterations.

To implement the initial idea of FWA that each firework in the swarm cooperate together to solve the optimization problem, it is needed to ensure that
the information for each firework is passed to the next iteration.
In the new framework of FWA, the independent selection operator will be performed for each firework, \textit{i.e.} each firework will select the best candidate among its all generated sparks and itself in each iteration respectively (cf. Alg.\ref{Alg:NEWFramework}).

\subsubsection{The Crowdness-avoiding Cooperative Strategy among the Fireworks}
In the fireworks swarm, each firework will generate a number of explosion sparks which can represent the quality of the local regions.
The quality of each firework's position is shared in the population to accelerate the convergence speed.

For the calculation of explosion amplitudes and explosion sparks number in CoFFWA, CF still takes the dynamic explosion amplitude strategy, while for the rest of fireworks, the explosion amplitudes are calculated as in dynFWA by Eq.(\ref{Eq:Amplitude}). For explosion spark number calculation, it is same in CoFFWA as in dynFWA as well by Eq.(\ref{Eq:sonnum}). After the generating of explosion sparks, each firework performs the independent selection method, respectively.

Moreover, In COFFWA, a crowdness-avoiding operation is introduced in the fireworks population.
For the crowdness-avoiding operation, it means that whenever any fireworks in the fireworks swarm is close to CF, \textit{i.e.} within a fixed range of CF, then the position of this firework will be reinitialized in the feasible search space.
The following gives two main reasons for the crowdness-avoiding operation:
\begin{itemize}
  \item The selected firework ($X_i$) has worse fitness and smaller explosion sparks number than the CF, thus the region that $X_i$ locates is not promising, it is seen as a waste of  evaluation times to continue the search for the fireworks with worse fitness around the CF.
  \item To increase the diversity of the fireworks swarm, it is better that the fireworks are not located close to each other.
\end{itemize}

\begin{figure}
\centering
  \includegraphics[width=0.3\textwidth]{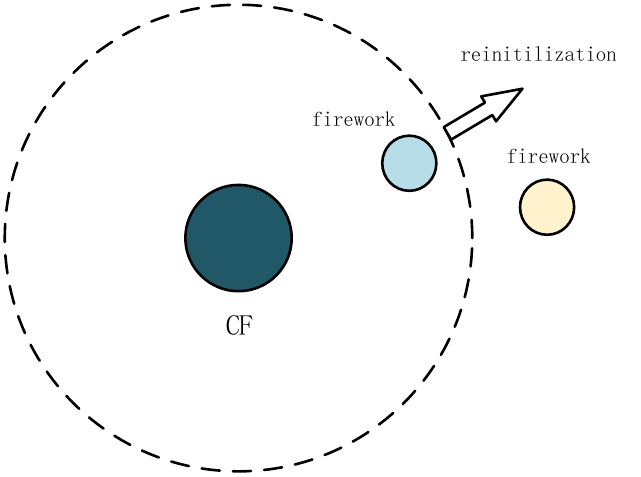}
  \caption{The Crowdness-avoiding Cooperative Strategy.}\label{Fig:CoFFWA}
\end{figure}

\begin{algorithm}[ht]
\caption{-- The crowdness-avoiding cooperative strategy}
\label{Alg:simple}
\begin{algorithmic}[1]
	\IF{$||X_i-X_{CF}||_\infty < \tau$}
    \STATE{Reinitialize the position of $X_i$}
    \ENDIF
\end{algorithmic}
\end{algorithm}

\section{Experiments Design}
\label{Sec:ExperimentsDesign}
To validate the ideas presented previously, several groups of experiments are designed.

\subsection{Significance Analysis of non-CFs/non-GCFs in Explosion Operator}

To investigate whether the non-CFs/non-GCFs are effective or not compared with the CF/GCF in the explosion operator, the following evaluation criterions, \textit{i.e.} significant improvement and resource cost are designed.

For these following criterions evaluation, FWA variants without Gaussian mutation operator, \textit{i.e.} the algorithms only generate explosion sparks are used to offset the influences of Gaussian mutation operators.

%
\subsubsection{Significant Improvement}
Among the fireworks swarm, if a firework $X_i$ generates the explosion spark with minimal fitness among all the explosion sparks and fireworks, then $X_i$ is seen to have one time significant improvement to the optimization. In the optimization process of FWA, at each iteration, at most one firework can make the significant improvement, thus to compare the performance of CF, GCF and non-CFs, non-GCFs, the significant improvement times of them during each run are recorded.

\begin{itemize}
  \item $\alpha_{CF}$, the percentage of CF's significant improvement times among all the significant improvement times.
  \item $\beta_{CF}$, the percentage of CF's significant improvement times among all the significant improvement times recorded from the $\frac{1}{30}*{E_{\mathrm{max}}}$-th evaluation time.
  \item $\alpha_{GCF}$, the percentage of GCF's significant improvement times among all the significant improvement times.
  \item $\beta_{GCF}$, the percentage of GCF's significant improvement times among all the significant improvement times recorded from the $\frac{1}{30}*{E_{\mathrm{max}}}$-th evaluation time.
\end{itemize}
Here, $E_{max}$ denotes the max number of evaluation times. It is thought that $\beta_{CF}$/$\beta_{GCF}$ are better than $\alpha_{CF}$/$\alpha_{GCF}$ for the performance comparison as at the early phase of the optimization, it is
likely that non-CFs gain more significant improvement times due to the great number of explosion sparks. However, significant improvements in the later phase of the optimization
are more important.
\subsubsection{Resources Cost}
For the optimization, the evaluation times ($E_{\mathrm{max}}$) is usually set to $D\times 10\,000$, where $D$ is the dimension of the problem~\cite{liang2013problem}.
\begin{itemize}
  \item $\theta_{CF}$, the percentage of CF's evaluation times among all the evaluation times.
  \item $\theta_{GCF}$, the percentage of GCF's evaluation times among all the evaluation times.
\end{itemize}

\subsection{Significance Analysis of Gaussian Mutation Operator}
To validate whether the Gaussian mutation operator introduced in EFWA is effective or not, the following comparisons are made.
\begin{itemize}
  \item EFWA-G vs EFWA-NG.
  \item dynFWA-G vs dynFWA-NG.
  \item AFWA-G vs AFWA-NG,
\end{itemize}
here, ``G'' and ``NG'' refer to with and without Gaussian mutation operator, respectively.

\subsection{Significance Analysis of the Proposed CoFFWA}
To validate the performance of the proposed CoFFWA, experiments are conducted to compare the performance of CoFFWA
with
EFWA~\cite{zhengCEC2013},
dynFWA~\cite{DynFWA},
AFWA~\cite{AFWA},
artificial bee colony (ABC)~\cite{el2013testing,karaboga2005idea},
differential evolution (DE)~\cite{padhye2013differential},
standard PSO in 2007 (SPSO2007)\cite{bratton2007},
and the most recent SPSO in 2011 (SPSO2011)~\cite{spso2011paper}.

\subsection{Experimental Setup and Platform}
Similar to dynFWA, the number of fireworks in CoFFWA is set to 5 and the explosion sparks number is set to 150. The reduction and amplification factors $C_r$ and $C_a$ are also empirically set to $0.9$ and $1.2$, respectively. For the parameter $\tau$  is set to 10*$A_{CF}$, and the maximum explosion amplitude for CF is bounded with the search range. For the rest of parameters in CoFFWA, they are identical to dynFWA~\cite{DynFWA}. The parameters for AFWA, dynFWA, DE, ABC, SPSO2007 and SPSO2011 are
listed in~\cite{AFWA}~\cite{karaboga2005idea}~\cite{DynFWA}~\cite{padhye2013differential}~\cite{bratton2007}~\cite{SPSO2011}, respectively.

In the experiments, the CEC2013 benchmark functions with 28 functions are used as the test suite and dimension is set to 30~\cite{liang2013problem}. For the recorded results in each algorithm, the number of runs is set to 51, and Wilcoxon signed-rank test is used to validate the performance improvement.
The experimental platform is MATLAB 2011b (Windows 7; Intel Core i7-2600 CPU @ 3.7 GHZ; 8 GB RAM).

\section{Experimental Results}
\label{Sec:ExperimentalResults}
\subsection{Significance Analysis of non-CFs/non-GCFs in Explosion Operator}

\begin{figure*}
\subfigure[Case 1: EFWA-NG, CF]{
  \includegraphics[width=0.5\textwidth]{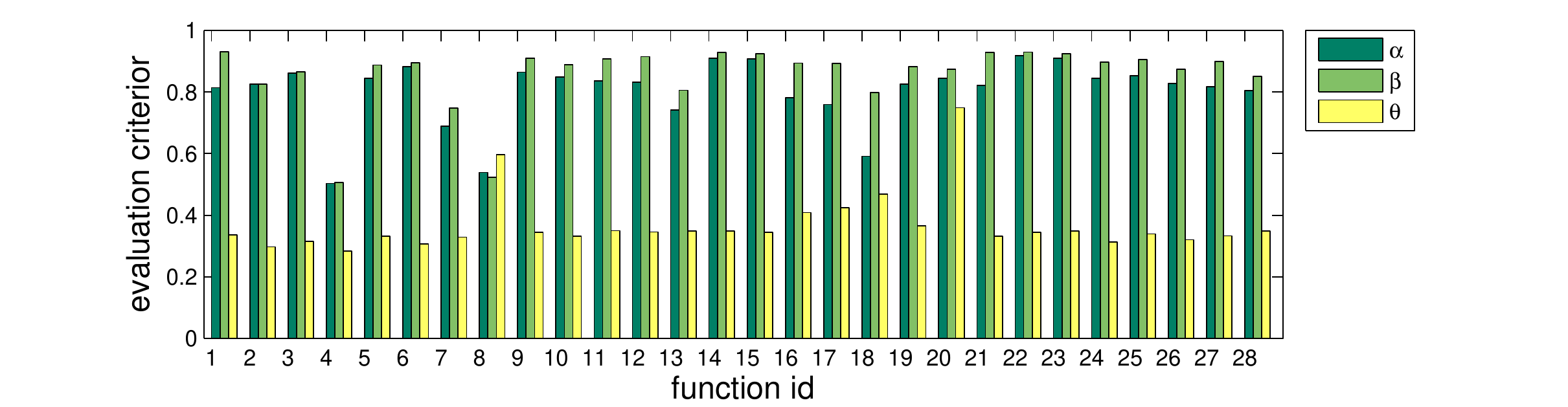}
  \label{Fig:C1}}
  \subfigure[Case 2: EFWA-NG, GCF]{
  \includegraphics[width=0.5\textwidth]{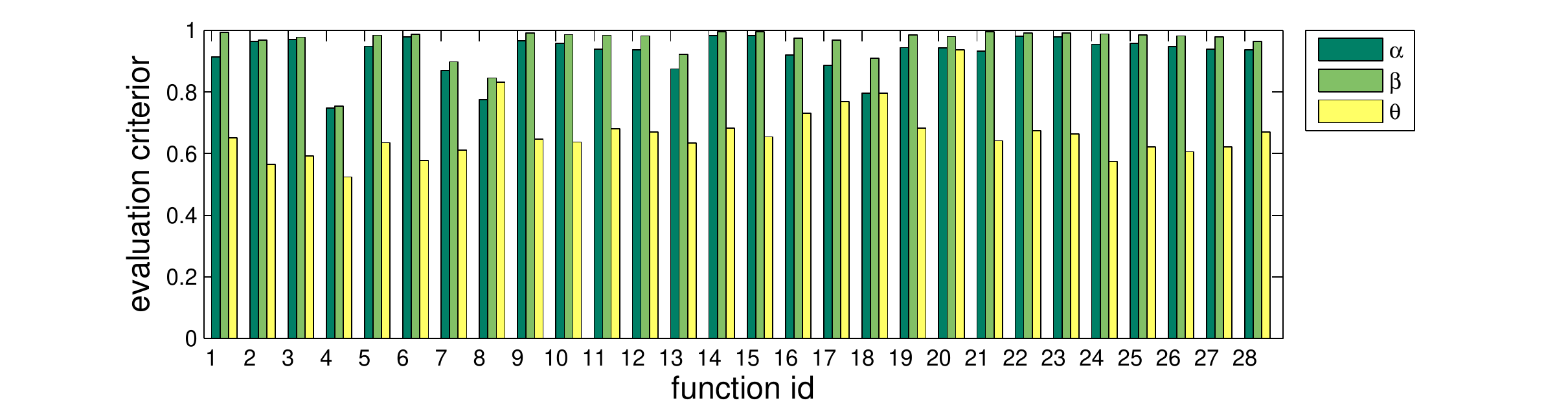}
  \label{Fig:C2}}\\
  \subfigure[Case 3: dynFWA-NG, CF]{
  \includegraphics[width=0.5\textwidth]{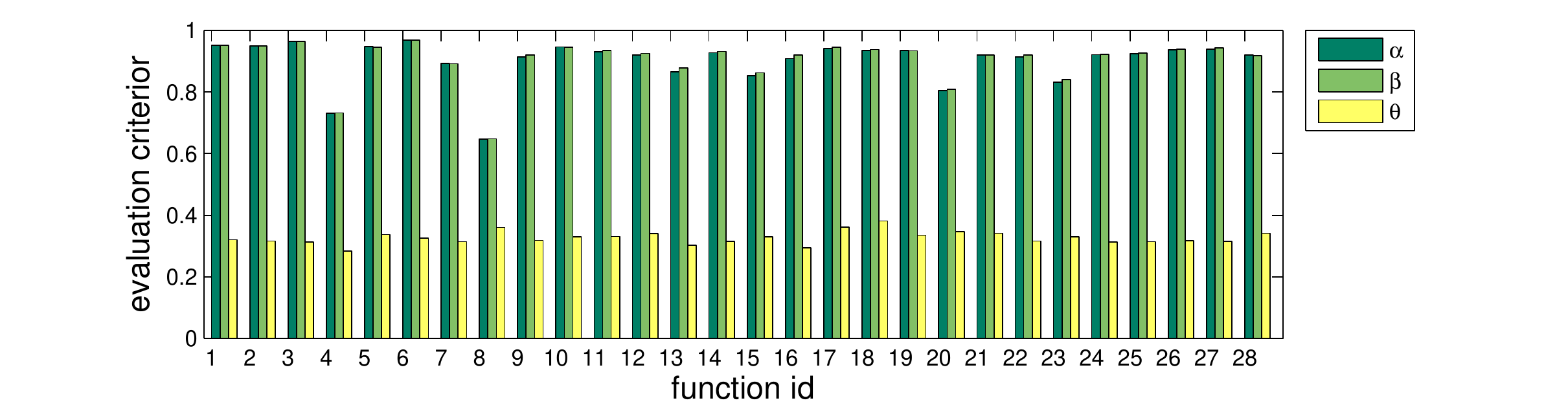}
  \label{Fig:C3}}
  \subfigure[Case 4: dynFWA-NG, GCF]{
  \includegraphics[width=0.5\textwidth]{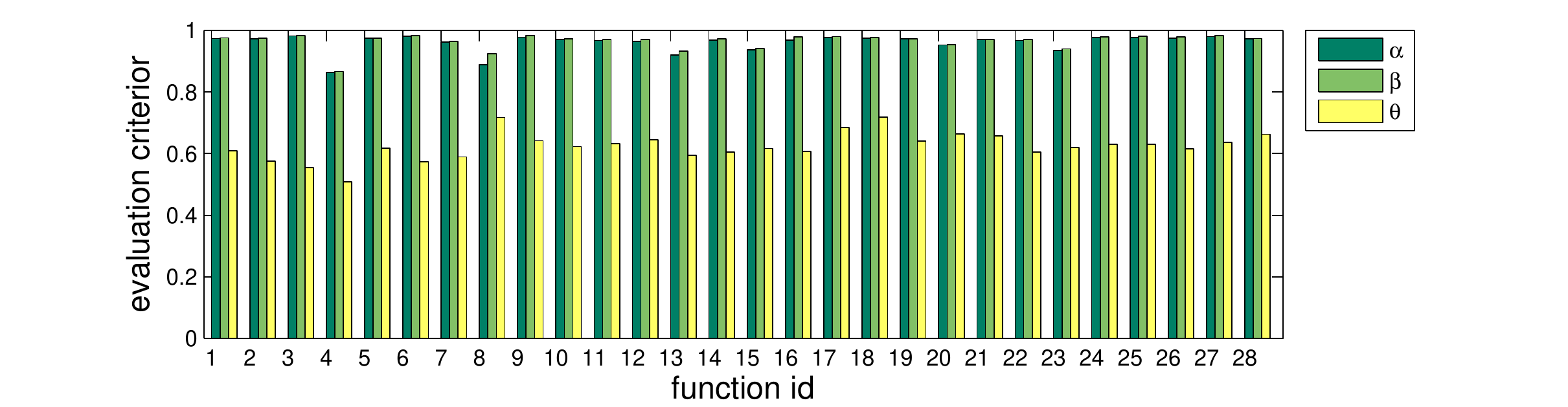}
  \label{Fig:C4}}
  \caption{Significant improvement and resources cost results of CF and GCF.}\label{Fig:Cases}
\end{figure*}

To validate the performance of CF and GCF, the $\alpha_{CF}$, $\beta_{CF}$, $\theta_{CF}$ and $\alpha_{GCF}$, $\beta_{GCF}$, $\theta_{GCF}$ are calculated on EFWA-NG and dynFWA-NG respectively, and the recorded results on 28 functions can be found in Fig.~\ref{Fig:Cases}.

Compare the performance between CF and non-CFs, it can be seen that, for both EFWA-NG and dynFWA-NG, CF takes smaller percentage of resources (\textit{i.e.} evaluation times), but makes the more significant improvement times to the search.

Compare the evaluation criterion $\alpha_{CF}$ and $\beta_{CF}$, it can be seen that for all functions, $\beta_{CF}$ which records the statistical results from the $\frac{1}{30}*E_{\mathrm{max}}$-th evaluation times is higher than $\alpha_{CF}$, which means that at the beginning of the optimization, the non-CFs have high probability for making significant improvements, while at the later phase, the chance goes smaller.
For GCF, the above situation is similar to CF.

Compare the results of CF with GCF, the GCF makes more significant improvements while taking more resources (\textit{i.e} evaluation times), due to that some fireworks except for CF may be close to CF, thus to have a high chance to make improvements.

In all, from the results in Fig.~\ref{Fig:Cases}, it can be concluded that the CF makes much more contributions than non-CF while taking smaller resources. While comparing the results of CF and GCF, it seems that the non-GCFs do not make any contribution to the search.

\subsection{Significance Analysis of Gaussian Mutation Operator}
Tab.~\ref{Tab:GaussianEvaluation} gives the experimental results of the versions of EFWA, dynFWA, AFWA with and without Gaussian mutation operator.
Comparing EFWA with EFWA-NG, it can be seen that EFWA-G is significantly better than EFWA-NG on 4 functions, which suggests that Gaussian mutation operator is effective to improve the performance for EFWA.
However, for dynFWA-NG and AFWA-NG, the versions without Gaussian mutation operator have better performance.
Moreover, in terms of the computational complexity, our previous results in \cite{DynFWA} suggest that Gaussian mutation operator is more time consuming than explosion sparks operator.
Thus, the Gaussian mutation operator will be removed from CoFFWA.

\begin{table}[!htbp]
  \centering
  \caption{Wilcoxon signed-rank test results for EFWA-G $vs$ EFWA-NG and dynFWA-G $vs$ dynFWA-NG and AFWA-G $vs$ AFWA-NG (bold values indicate the performance difference is significant, while 1/0/-1 denotes the version with Gaussian sparks operator is significant better / not significant different / significant worse than the version without Gaussian sparks.).}
  \resizebox{0.95\linewidth}{!}{
    \begin{tabular}{c|cc|cc|cc}
    \hline
    \multirow{2}{*}{F.}&EFWA-G $vs$ &&dynFWA-G $vs$ &&AFWA-G $vs$& \\
   & EFWA-NG && dynFWA-NG && AFWA-NG& \\
    \hline \hline
    1     & \textbf{2.316E-03} & 1 & 1.000E+00 & 0     & 1.000E+00 & 0 \\
    2     & 4.256E-01 & 0     & 9.328E-01 & 0     & 4.647E-01 & 0 \\
    3     & 8.956E-01 & 0     & 2.339E-01 & 0     & 7.191E-02 & 0 \\
    4     & 7.858E-01 & 0     & 7.492E-02 & 0     & \textbf{5.689E-03} & -1 \\
    5     & \textbf{4.290E-02} & 1 & 7.646E-02 & 0     & 5.239E-01 & 0 \\
    6     & 1.654E-01 & 0     & 7.858E-01 & 0     & 9.030E-01 & 0 \\
    7     & 9.552E-01 & 0     & 6.869E-01 & 0     & 2.728E-01 & 0 \\
    8     & 9.776E-01 & 0     & 4.704E-01 & 0     & 8.808E-01 & 0 \\
    9     & 5.178E-01 & 0     & 4.997E-01 & 0     & 7.571E-01 & 0 \\
    10    & 3.732E-01 & 0     & 5.057E-01 & 0     & \textbf{9.545E-03} & -1 \\
    11    & 5.830E-02 & 0     & 6.629E-01 & 0     & 3.204E-01 & 0 \\
    12    & 6.193E-01 & 0     & 3.783E-01 & 0     & 1.801E-01 & 0 \\
    13    & 8.220E-01 & 0     & 1.863E-01 & 0     & 4.590E-01 & 0 \\
    14    & 4.101E-02 & 0     & 3.834E-01 & 0     & 5.239E-01 & 0 \\
    15    & 6.869E-01 & 0     & 3.438E-01 & 0     & 4.879E-01 & 0 \\
    16    & 2.811E-01 & 0     & 4.256E-01 & 0     & 8.220E-01 & 0 \\
    17    & 9.179E-01 & 0     & 1.863E-01 & 0     & 2.339E-01 & 0 \\
    18    & 6.938E-01 & 0     & 4.762E-01 & 0     & 6.460E-01 & 0 \\
    19    & 9.402E-01 & 0     & 1.542E-01 & 0     & 7.786E-01 & 0 \\
    20    & \textbf{1.559E-02} & -1     & 5.830E-02 & 0     & 4.997E-01 & 0 \\
    21    & \textbf{6.910E-04} & 1     & 7.997E-01 & 0     & 5.937E-01 & 0 \\
    22    & 9.776E-01 & 0     & 3.583E-01 & 0     & 4.202E-01 & 0 \\
    23    & 7.217E-01 & 0     & 7.642E-01 & 0     & 6.260E-01 & 0 \\
    24    & \textbf{1.079E-02} & 1     & 5.486E-01 & 0     & 7.500E-01 & 0 \\
    25    & 8.734E-01 & 0     & 2.091E-01 & 0     & \textbf{1.369E-02} & -1 \\
    26    & 2.687E-01 & 0     & 7.217E-01 & 0     & 3.834E-01 & 0 \\
    27    & 3.534E-01 & 0     & 8.734E-01 & 0     & 1.597E-01 & 0 \\
    28    & 6.460E-01 & 0     & \textbf{0.000E+00} & -1    & \textbf{3.285E-03} & -1 \\
    \bottomrule
    \end{tabular}}
  \label{Tab:GaussianEvaluation}%
\end{table}%

\subsection{Significance Analysis of CoFFWA}
Tab.~\ref{Tab:MeanRank} shows the mean fitness value over 51 runs for the 28 functions for
 ABC, DE, SPSO2007, SPSO2011, EFWA, AFWA, dynFWA and CoFFWA and the corresponding rank of each algorithm.
It can be seen that CoFFWA is able to achieve better results than dynFWA(AFWA) on 24(22) functions while dynFWA(AFWA) is better than CoFFWA on 3(5) functions. For 1(1) function, the results are identical.
For the comparison with EFWA, the advantage is especially obvious.

The Wilcoxon signed-rank test results in Tab.~\ref{Tab:ttest} suggest that CoFFWA is significantly better than dynFWA(AFWA) on 8(8) functions while significant worse on 2(2) functions. Furthermore, in terms of computation complexity (cf Tab.~\ref{Tab:runtime}), CoFFWA has the similar computation cost as dynFWA.
Runtime results in Tab.~\ref{Tab:runtime} suggest that
compared to EFWA, dynFWA, the proposed CoFFWA has almost the same
runtime.

Compared with ABC, DE, SPSO2007, SPSO2011, EFWA, AFWA and dynFWA in term of mean fitness rank, it can be seen that CoFFWA achieves the best optimization results and performances, \textit{i.e.} the minimal average of mean fitness rank (3.00 in Tab.~\ref{Tab:MeanRank}), thus we can conclude that the proposed cooperative framework for FWA is significant.

\begin{table*}[htbp]
  \centering
  \caption{Mean Fitness and mean fitness rank of ABC, DE, SPSO2007, SPSO2011, EFWA, AFWA, dynFWA and CoFFWA (AR denotes the average of mean rank value).}
  \label{Tab:MeanRank}
    \resizebox{1\linewidth}{!}{
    \begin{tabular}{l|rr|rr|rr|rr|rr|rr|rr|rr}
    \hline
    F     & ABC   &       & DE    &       & SPSO2007 &       & SPSO2011 &       & EFWA  &       & AFWA  &       & dynFWA &       & CoFFWA &  \\
    \hline \hline
  1     & 0.00E+00 & 1     & 1.89E-03 & 7     & 0.00E+00 & 1     & 0.00E+00 & 1     & 8.50E-02 & 8     & 0.00E+00 & 1     & 0.00E+00 & 1     & 0.00E+00 & 1 \\
    2     & 6.20E+06 & 8     & 5.52E+04 & 1     & 6.08E+06 & 7     & 3.38E+05 & 2     & 5.85E+05 & 3     & 8.92E+05 & 6     & 8.71E+05 & 4     & 8.80E+05 & 5 \\
    3     & 5.74E+08 & 7     & 2.16E+06 & 1     & 6.63E+08 & 8     & 2.88E+08 & 6     & 1.16E+08 & 3     & 1.26E+08 & 5     & 1.23E+08 & 4     & 8.04E+07 & 2 \\
    4     & 8.75E+04 & 7     & 1.32E-01 & 1     & 1.03E+05 & 8     & 3.86E+04 & 6     & 1.22E+00 & 2     & 1.14E+01 & 4     & 1.04E+01 & 3     & 2.01E+03 & 5 \\
    5     & 0.00E+00 & 1     & 2.48E-03 & 7     & 0.00E+00 & 2     & 5.42E-04 & 3     & 8.05E-02 & 8     & 6.00E-04 & 5     & 5.51E-04 & 4     & 7.41E-04 & 6 \\
    6     & 1.46E+01 & 2     & 7.82E+00 & 1     & 2.52E+01 & 4     & 3.79E+01 & 8     & 3.22E+01 & 7     & 2.99E+01 & 5     & 3.01E+01 & 6     & 2.47E+01 & 3 \\
    7     & 1.25E+02 & 7     & 4.89E+01 & 1     & 1.13E+02 & 6     & 8.79E+01 & 2     & 1.44E+02 & 8     & 9.19E+01 & 4     & 9.99E+01 & 5     & 8.99E+01 & 3 \\
    8     & 2.09E+01 & 6     & 2.09E+01 & 2     & 2.10E+01 & 7     & 2.09E+01 & 5     & 2.10E+01 & 8     & 2.09E+01 & 4     & 2.09E+01 & 3     & 2.09E+01 & 1 \\
    9     & 3.01E+01 & 8     & 1.59E+01 & 1     & 2.93E+01 & 6     & 2.88E+01 & 5     & 2.98E+01 & 7     & 2.48E+01 & 4     & 2.41E+01 & 3     & 2.40E+01 & 2 \\
    10    & 2.27E-01 & 5     & 3.24E-02 & 1     & 2.38E-01 & 6     & 3.40E-01 & 7     & 8.48E-01 & 8     & 4.73E-02 & 3     & 4.81E-02 & 4     & 4.10E-02 & 2 \\
    11    & 0.00E+00 & 1     & 7.88E+01 & 3     & 6.26E+01 & 2     & 1.05E+02 & 7     & 2.79E+02 & 8     & 1.05E+02 & 6     & 1.04E+02 & 5     & 9.90E+01 & 4 \\
    12    & 3.19E+02 & 7     & 8.14E+01 & 1     & 1.15E+02 & 3     & 1.04E+02 & 2     & 4.06E+02 & 8     & 1.52E+02 & 5     & 1.58E+02 & 6     & 1.40E+02 & 4 \\
    13    & 3.29E+02 & 7     & 1.61E+02 & 1     & 1.79E+02 & 2     & 1.94E+02 & 3     & 3.51E+02 & 8     & 2.36E+02 & 4     & 2.54E+02 & 6     & 2.50E+02 & 5 \\
    14    & 3.58E-01 & 1     & 2.38E+03 & 3     & 1.59E+03 & 2     & 3.99E+03 & 7     & 4.02E+03 & 8     & 2.97E+03 & 5     & 3.02E+03 & 6     & 2.70E+03 & 4 \\
    15    & 3.88E+03 & 4     & 5.19E+03 & 8     & 4.31E+03 & 7     & 3.81E+03 & 3     & 4.28E+03 & 6     & 3.81E+03 & 2     & 3.92E+03 & 5     & 3.37E+03 & 1 \\
    16    & 1.07E+00 & 5     & 1.97E+00 & 8     & 1.27E+00 & 6     & 1.31E+00 & 7     & 5.75E-01 & 3     & 4.97E-01 & 2     & 5.80E-01 & 4     & 4.56E-01 & 1 \\
    17    & 3.04E+01 & 1     & 9.29E+01 & 2     & 9.98E+01 & 3     & 1.16E+02 & 5     & 2.17E+02 & 8     & 1.45E+02 & 7     & 1.43E+02 & 6     & 1.10E+02 & 4 \\
    18    & 3.04E+02 & 8     & 2.34E+02 & 7     & 1.80E+02 & 4     & 1.21E+02 & 1     & 1.72E+02 & 2     & 1.75E+02 & 3     & 1.88E+02 & 6     & 1.80E+02 & 5 \\
    19    & 2.62E-01 & 1     & 4.51E+00 & 2     & 6.48E+00 & 3     & 9.51E+00 & 7     & 1.24E+01 & 8     & 6.92E+00 & 5     & 7.26E+00 & 6     & 6.51E+00 & 4 \\
    20    & 1.44E+01 & 6     & 1.43E+01 & 5     & 1.50E+01 & 8     & 1.35E+01 & 4     & 1.45E+01 & 7     & 1.30E+01 & 1     & 1.33E+01 & 3     & 1.32E+01 & 2 \\
    21    & 1.65E+02 & 1     & 3.20E+02 & 6     & 3.35E+02 & 8     & 3.09E+02 & 3     & 3.28E+02 & 7     & 3.16E+02 & 5     & 3.10E+02 & 4     & 2.06E+02 & 2 \\
    22    & 2.41E+01 & 1     & 1.72E+03 & 2     & 2.98E+03 & 3     & 4.30E+03 & 7     & 5.15E+03 & 8     & 3.45E+03 & 6     & 3.33E+03 & 5     & 3.32E+03 & 4 \\
    23    & 4.95E+03 & 5     & 5.28E+03 & 6     & 6.97E+03 & 8     & 4.83E+03 & 4     & 5.73E+03 & 7     & 4.70E+03 & 2     & 4.75E+03 & 3     & 4.47E+03 & 1 \\
    24    & 2.90E+02 & 7     & 2.47E+02 & 1     & 2.90E+02 & 6     & 2.67E+02 & 2     & 3.05E+02 & 8     & 2.70E+02 & 4     & 2.73E+02 & 5     & 2.68E+02 & 3 \\
    25    & 3.06E+02 & 6     & 2.80E+02 & 1     & 3.10E+02 & 7     & 2.99E+02 & 5     & 3.38E+02 & 8     & 2.99E+02 & 4     & 2.97E+02 & 3     & 2.94E+02 & 2 \\
    26    & 2.01E+02 & 1     & 2.52E+02 & 3     & 2.57E+02 & 4     & 2.86E+02 & 7     & 3.02E+02 & 8     & 2.73E+02 & 6     & 2.61E+02 & 5     & 2.13E+02 & 2 \\
    27    & 4.16E+02 & 1     & 7.64E+02 & 2     & 8.16E+02 & 3     & 1.00E+03 & 7     & 1.22E+03 & 8     & 9.72E+02 & 5     & 9.80E+02 & 6     & 8.71E+02 & 4 \\
    28    & 2.58E+02 & 1     & 4.02E+02 & 5     & 6.92E+02 & 7     & 4.01E+02 & 4     & 1.23E+03 & 8     & 4.37E+02 & 6     & 2.96E+02 & 3     & 2.84E+02 & 2 \\
    &   \multicolumn{2}{c}{AR: 4.14 }&     \multicolumn{2}{c}{AR: 3.18} &     \multicolumn{2}{c}{AR: 5.04} &     \multicolumn{2}{c}{AR:  4.64} &    \multicolumn{2}{c}{AR: 6.79}&      \multicolumn{2}{c}{ AR: 4.25} &    \multicolumn{2}{c}{ AR: 4.43} &     \multicolumn{2}{c}{ \textbf{AR: 3.00}} \\
	\hline
    \end{tabular}}
\end{table*}%

\begin{table}[!htbp]
  \centering
  \caption{Wilcoxon signed-rank test results for CoFFWA $vs$ AFWA and CoFFWA $vs$ dynFWA ( 1/0/-1 denotes the CoFFWA is significant better / not significant different / significant worse than EFWA/AFWA/dynFWA).}
  \label{Tab:ttest}%
  \resizebox{0.68\linewidth}{!}{
    \begin{tabular}{c|cc|cc}
    \hline
    \multirow{2}{*}{F.}&CoFFWA $vs$ &&CoFFWA $vs$& \\
   & AFWA && dynFWA& \\
    \hline \hline
     1     & 1.000E+00 & 0     & 1.000E+00 & 0 \\
    2     & 6.938E-01 & 0     & 6.460E-01 & 0 \\
    3     & 1.313E-01 & 0     & 9.328E-01 & 0 \\
    4     & 0.000E+00 & -1    & 0.000E+00 & -1 \\
    5     & 0.000E+00 & -1    & 0.000E+00 & -1 \\
    6     & 2.160E-01 & 0     & 3.989E-01 & 0 \\
    7     & 1.132E-01 & 0     & 3.889E-03 & 1 \\
    8     & 5.802E-01 & 0     & 7.077E-01 & 0 \\
    9     & 4.202E-01 & 0     & 9.850E-01 & 0 \\
    10    & 2.332E-01 & 0     & 2.981E-01 & 0 \\
    11    & 2.853E-01 & 0     & 2.413E-01 & 0 \\
    12    & 2.528E-01 & 0     & 7.963E-02 & 0 \\
    13    & 4.041E-01 & 0     & 5.867E-01 & 0 \\
    14    & 1.334E-02 & 1     & 8.676E-03 & 1 \\
    15    & 8.600E-05 & 1     & 8.900E-05 & 1 \\
    16    & 3.486E-01 & 0     & 1.153E-01 & 0 \\
    17    & 1.000E-06 & 1     & 3.000E-06 & 1 \\
    18    & 2.091E-01 & 0     & 4.148E-01 & 0 \\
    19    & 4.094E-01 & 0     & 2.567E-01 & 0 \\
    20    & 1.132E-01 & 0     & 6.260E-01 & 0 \\
    21    & 0.000E+00 & 1     & 1.000E-06 & 1 \\
    22    & 2.489E-01 & 0     & 8.513E-01 & 0 \\
    23    & 2.811E-01 & 0     & 1.197E-01 & 0 \\
    24    & 8.440E-01 & 0     & 2.528E-01 & 0 \\
    25    & 3.659E-02 & 1     & 3.783E-01 & 0 \\
    26    & 3.300E-05 & 1     & 2.316E-03 & 1 \\
    27    & 2.966E-02 & 1     & 5.528E-03 & 1 \\
    28    & 2.865E-02 & 1     & 1.594E-03 & 1 \\
    & \multicolumn{2}{c}{8+/18/2-}&       \multicolumn{2}{c}{ 8+/18/2-}\\
    \hline
    \end{tabular}}
\end{table}%

\begin{table}[t]
  \centering
  \caption{Runtime Comparison}
  \resizebox{1\linewidth}{!}{
\begin{tabular}{c|c||c|c||c|c||c|c}
  \hline
  EFWA & 1.3 & AFWA & --- & dynFWA & 1 & CoFFWA & 1.05\\  \hline
\end{tabular}}
  \label{Tab:runtime}%
\end{table}%



\section{Conclusion}
\label{Sec:futurework}
In this paper, we have presented a cooperative framework for FWA (CoFFWA).
The contributions include three aspects.
\begin{enumerate}
  \item The cooperative strategies in conventional FWA are analyzed and evaluated. In the explosion operator, the performance of CF, GCF, non-CFs and non-GCFs in EFWA, dynFWA and AFWA are investigated and the designed criterions are recorded to support our opinion that the non-CF/non-GCFs do not make much contribution to the optimization for a specific problem while taking a lot of evaluation times. For the Gaussian mutation operator, experimental results suggest that it is not as effective as it is designed to be.
  \item Based on the analysis of cooperative strategies in conventional FWA framework, a cooperative framework of FWA with an independent selection method and crowdness-avoiding cooperative strategy is finally proposed  to ensure the information inheritance and improve the diversity of fireworks population.
  \item Moreover, the explosion amplitude strategies in AFWA and dynFWA are compared and the relationship between them is presented.
\end{enumerate}

Compared with PSO's search manner--- \textit{one individual generates one individual}, CoFFWA presents a cooperative explosive search manner ---  \textit{one individual generates a number of individuals} which has the ability to estimate the property of typical local search space and will make the search direction of the swarm stable and the cooperative strategies among the fireworks make the firework with better fitness will have more resources.
In the future, we need to focus on this unique search manner and try to develop new kinds of cooperative search strategies among the fireworks swarm to enhance the exploration and exploitation capacities.

\ifCLASSOPTIONcompsoc
  \section*{Acknowledgments}
\else
  \section*{Acknowledgment}
\fi

This work was supported by the Natural Science Foundation
of China (NSFC) under grant no. 61375119 and 61170057,
and partially supported by National Key Basic Research Development
Plan (973 Plan) Project of China with grant no.
2015CB352302. Prof. Ying Tan is the corresponding author.

\ifCLASSOPTIONcaptionsoff
  \newpage
\fi



%
%
%

\bibliographystyle{IEEEtran}
\bibliography{bare_adv}
\end{document}